\documentclass{article}
\usepackage{ijcai25}
\usepackage[subtle]{savetrees}

\usepackage{times}
\usepackage{soul}
\usepackage{url}
\usepackage[hidelinks]{hyperref}
\usepackage[utf8]{inputenc}
\usepackage[small]{caption}
\usepackage{graphicx}
\usepackage{amsmath}
\usepackage{amsthm}
\usepackage{booktabs}
\usepackage{algorithm}
\usepackage{algorithmic}
\usepackage[switch]{lineno}

\usepackage{bbm}
\usepackage{amssymb}
\usepackage{natbib}

\newcommand{\D}{\mathcal{D}}

\newcommand{\N}{\mathcal{N}}
\newcommand{\E}{\mathbb{E}}
\newcommand{\bs}{\boldsymbol}
\newcommand{\bx}{\textbf{x}}
\newcommand{\bX}{\textbf{X}}

\newcommand{\bY}{\textbf{Y}}
\newcommand{\bW}{\textbf{W}}

\newcommand{\bV}{\textbf{V}}
\newcommand{\bc}{\textbf{c}}
\newcommand{\bff}{\textbf{f}}

\newcommand{\bT}{\textbf{T}}
\newcommand{\bp}{\textbf{p}}

\newcommand{\bu}{\textbf{u}}
\newcommand{\bd}{\textbf{d}}

\newcommand{\tY}{\tilde{\bY}}
\newcommand{\tV}{\tilde{\bV}}

\newcommand{\GP}{\mathcal{G}\mathcal{P}}

\usepackage{multirow}
\usepackage{subfigure}

\title{Diffusion-aware Censored Gaussian Processes for Demand Modelling}

\author{
    Filipe Rodrigues
    \affiliations
    Technical University of Denmark
    \emails
    rodr@dtu.dk
}

\begin{document}

\maketitle

\begin{abstract}
Inferring the true demand for a product or a service from aggregate data is often challenging due to the limited available supply, thus resulting in observations that are censored and correspond to the realized demand, thereby not accounting for the unsatisfied demand. Censored regression models are able to account for the effect of censoring due to the limited supply, but they don't consider the effect of substitutions, which may cause the demand for similar alternative products or services to increase. This paper proposes Diffusion-aware Censored Demand Models, which combine a Tobit likelihood with a graph diffusion process in order to model the latent process of transfer of unsatisfied demand between similar products or services. We instantiate this new class of models under the framework of GPs and, based on both simulated and real-world data for modeling sales, bike-sharing demand, and EV charging demand, demonstrate its ability to better recover the true demand and produce more accurate out-of-sample predictions. 
\end{abstract}

\section{Introduction}

Learning well-specified probabilistic models capable of dealing with censored data is a long-standing challenge of the statistical sciences and machine learning. Censoring occurs when the value of a given observation or measurement is only partially known, with the true value being \emph{latent}. Censoring is a process that occurs naturally in many research fields, such as economics, natural sciences, and medicine \citep{breen1996regression}. In this paper, we are particularly interested in censoring in the context of demand modeling. Demand modeling is crucial across various application domains, such as retail and e-commerce, energy, transportation and logistics, healthcare, telecommunications, manufacturing, tourism, etc., as it enables accurate forecasting, resource optimization, and strategic decision-making by understanding and predicting consumer needs, market trends, and system requirements. When modeling the aggregate demand for a product or service, censoring occurs whenever it runs out of supply, thus resulting in observations that are upper-bounded by the available supply (\emph{satisfied demand}) and, therefore, don't correspond to the actual (\emph{true}) demand \citep{heien1990demand}, which is the quantity of interest to be modeled. From a learning perspective, if the dependent variable is censored for a non-neglectable fraction of the observations, then the parameter estimates obtained (e.g., by standard regression approaches such as OLS) will be inherently biased. 

In the statistics literature, Tobit regression models \citep{amemiya1984tobit,mcdonald1980uses} constitute the main workhorse for handling censored observations. They provide a framework capable of accounting for left- and right-censored data through a carefully designed likelihood function. However, they rely on the assumption that individual censoring processes occur in isolation, which is unrealistic for many real-world applications. Consider the problem of modeling the charging demand of electric vehicles (EVs) in an area. If, at certain periods of the day, all chargers are occupied, then the observed charging demand may not correspond to the true demand, which will likely be higher than the value observed. Critically though, that \emph{unsatisfied demand} above the available supply is not lost and, at least in a fraction of the cases, it will be transferred to the nearest available charger \citep{hipolito2022charging}, thus resulting in an unusually high demand being observed at that charging location. Besides leading to underestimation of the true demand in the original area, this may erroneously lead one to believe that the need for charging in nearby areas is higher than it actually is. More generally, we can consider the problem of modeling aggregate demand data for a given product or service. If the supply for that product reaches zero, then there is a high likelihood that that will result in a higher demand being observed for other similar/substitute products \citep{wan2018demand} - e.g., supermarket customers buying more rucola because it has run out of lettuce to sell, or simply buying from a different brand.  %Moreover, it should be possible to leverage the information about the unusually increased observed in one area to better infer the true demand for the nearby areas being censored by the limited supply. 

This paper aims to model the ubiquitous process of demand transfer between related products or services due to censoring from an aggregate (macro-economic) perspective. It proposes a new class of models, which we refer to as Diffusion-aware Censored Demand Models, that integrates ideas from Tobit models \citep{tobin1958estimation,amemiya1984tobit} with a graph diffusion process to model the latent process of transfer of unsatisfied demand. In doing so, we provide a statistically sound methodology to more accurately infer the latent true demand for the product subject to censoring by taking into account the unusually high demand for similar products, as well as to obtain adjusted estimates for the demand of the latter by accounting for the ``spillover effect''. By modeling the demand transfer process that occurs between similar products or services when their observed demand is subject to censoring due to limited supply, we are able to obtain more unbiased regression models of the true aggregate demand. We instantiate this new class of models under the framework of Gaussian processes (GPs) and propose Diffusion-aware Censored GPs, thus allowing us to jointly model multiple correlated time-series of aggregate demand data corresponding to different products or services. We begin by empirically validating the proposed approach using artificial data under carefully constructed scenarios, whereby we analyze its strengths and limitations. We then demonstrate the proposed approach using three real-world datasets for modeling sales data, bike-sharing demand, and EV charging demand. Through these experiments, we pinpoint the advantages of the proposed approach over standard censored regression, thus demonstrating its ability to better recover the true demand and produce more accurate out-of-sample predictions. 

\section{Related work}

Censoring arises naturally in many research fields. Besides the cases in economics discussed above, it occurs, for example, in the natural sciences whenever the true value of interest is outside the measurable range of a measuring instrument. Similarly, in medicine, censoring arises, for example, when measuring the survival time in clinical trials, since one may know that the survival time surpasses the present moment, but one cannot know the total survival time. Due to the pervasiveness and, oftentimes, critical importance of censored variables, it is unsurprising that a significant portion of the literature is dedicated to them. Historically, learning well-specified models of censored data has always been an important focus of statistical research, with Tobit models constituting a unifying probabilistic approach for censored data. %, thus proving that if the dependent variable in a regression model is censored for a non-neglectable fraction of the observations, OLS estimation is inappropriate. Its core idea consists of introducing a composite likelihood function, where for observations not subjective to censoring, a ``standard'' likelihood is used (e.g., Gaussian), while for censored observations (e.g., above a censoring threshold, in the case of right-censoring), a special likelihood is used in order to account for all the possible values that the variable can take (e.g., the likelihood of all values above the censoring threshold). 
With the advances in machine learning, the limitations of Tobit models associated with its linearity assumptions have been relaxed through the use of non-linear models such as Gaussian processes \citep{basson2023variational,gammelli2020estimating} and Neural Networks \citep{huttel2023mind}. However, when using a Tobit likelihood, inference is no longer tractable. Therefore, a considerable portion of the literature is dedicated to developing approximate inference approaches (e.g., \cite{chib1992bayes,groot2012gaussian}). 

One domain in which censored data is prevalent is urban mobility, from which we draw several case studies. People want to travel from A to B at a given time using a given transportation mode, but they often face supply constraints that force them to, for example, opt for an alternative transportation mode, change their departure time, or not travel at all. %In all cases, the result is an incurred cost for the person (money, time, comfort, etc.). 
Therefore, modeling and inferring true demand from censored demand observations in space and time is of vital importance for optimizing transportation systems. For example, \cite{gammelli2020estimating} proposed an approach based on independent Gaussian processes and a Tobit likelihood to model the demand for shared mobility services at individual locations, thus resulting in more accurate forecasts of the latent true aggregate demand. Similarly, \cite{huttel2022modeling} proposed using neural networks and censored quantile regression models to model EV car-sharing usage. %, while in \cite{boe2023modelling}, a similar approach was applied for modelling the difference between censored and uncensored electric vehicle charging demand. 
\cite{xie2023censored} further combine the predictions of a censored GP model with an optimization procedure to efficiently allocate resources in bike-sharing systems. 

Although these models are able to account for the censoring induced by the limited supply and adjust their estimates accordingly, they ignore the latent process of transfer of unsatisfied demand that the censoring process generates. For example, in bike-sharing systems, there is a high likelihood that users go to the nearest hub if there are no bikes available at the current hub, thus resulting in an abnormally high demand for the other hub. Researchers in econometrics are well aware of the importance of accounting for \emph{substitution effects} and have proposed several approaches to model them \citep{wan2018demand,schaafsma2020substitution,domarchi2023electric}. However, their focus is on micro-econometric approaches that model individual choice behavior and, therefore, require individual choice data for calibration, which can be difficult and costly to obtain. On the other hand, macro-economic data of aggregate demand is typically readily available. Contrasting with the existing literature, this paper proposes a novel framework to model this demand transfer process from a macro-economic perspective, thus resulting in more unbiased regression models of the true (latent) aggregate demand. 
%However, approaches such as \citep{gammelli2020estimating,xie2023censored} completely ignore this, thus resulting in biased estimates and, more importantly, sub-optimal resource allocation procedures. Contrasting with the existing literature, this paper proposes a new framework to model this demand transfer process thus resulting in more unbiased regression models of the true (latent) aggregate demand. 

\section{Background: Censored Gaussian Processes}
%\label{subsec:cgp}
\subsection{Censored Gaussian Processes}
\label{subsec:cgp}

Given a dataset $\D = \{(\bx_n, y_n, c_n)\}_{n=1}^N$, which, besides the covariates $\bx_n$ and respective targets $y_n$, also contains a binary indicator variable $c_n$ of whether the $n^{\text{th}}$ observation is subject to {censoring}, the goal of censored regression \citep{demaris2004regression,greene2003econometric} is to learn a function of the \emph{latent} true (uncensored) value $f_n = f(\bx_n)$. When modeling demand data, we are particularly interested in the case of right-censoring, where $y_n$ is upper-bounded by a given threshold $u_n$ corresponding to the available supply, such that
\begin{align}
	y_n = 
	\begin{cases}
		f_n, \quad \text{if } f_n < u_n \\
		u_n, \quad \text{if } f_n \geq u_n
	\end{cases}. 
\end{align}
The censoring indicator variable $c_n$ can then be formally defined as $c_n = \mathbbm{1}(f_n \geq u_n)$.
Traditionally, when modeling scalar-valued censored data, a reasonable choice for an observation model is the well-known Tobit likelihood \citep{tobin1958estimation,greene2003econometric}, or a type-I Tobit model according to the taxonomy of \citep{amemiya1984tobit}. For right-censored data, the likelihood of $y_n$ under a censored Gaussian distribution is given by
\begin{align}
	p(y_n \mid f_n) = \N(y_n \mid f_n, \, \sigma^2)^{(1-c_n)} \, (1-\Phi(y_n \mid f_n, \, \sigma^2))^{c_n}, 
	\label{eq:tobit_lik}
\end{align}
where $\Phi$ is the Gaussian cumulative density function (CDF). Note that this likelihood can be generalized to other distributions such as Poisson or Negative Binomial as discussed, for example, in \citep{gammelli2022generalized}. Censored GP regression \citep{basson2023variational,groot2012gaussian} then proceeds by placing a GP prior on $f(\bx) \sim \GP(0, \kappa(\bx,\bx'))$, and performing Bayesian inference to compute the posterior distribution over the true (uncensored) values $\{f_n\}$ or the predictive distribution for a new test point $\bx_*$.

\section{Problem formulation}
\label{subsec:problem_formulation}

We consider multivariate time-series data comprising input-output pairs, $\{\bX^{(tp)} \in \mathbb{R}^{N_t \times N_p \times D}, \bY^{(tp)} \in \mathbb{R}^{N_t \times N_p}\}$, where $N_t$ denotes the number of temporal points, $N_p$ the number of products or services, and $D = 1 + D_p$ the input dimensions, with $D_p$ being the number of product features. As it is common in the GP literature, we let $\bX = \text{vec}(\bX^{(tp)}) \in \mathbb{R}^{N \times D}$, $\bY = \text{vec}(\bY^{(tp)}) \in \mathbb{R}^{N \times 1}$, where $N = N_t N_p$ and the operator $\text{vec}(\cdot)$ simply converts the data into vector form, while preserving the ordering by time first and then products. For convenience, we use $\bX_{n,k} = \bX^{(tp)}_{n,k}$ and $\bY_{n,k} = \bY^{(tp)}_{n,k}$ to index the aggregate demand data for product $k$ at time index $n$. Our goal is then to learn a random function $f: \mathbb{R}^D \rightarrow \mathbb{R}$, on which we place a zero-mean GP prior with a fully-factorized likelihood, thus leading to the following generative process: 
\begin{align}
	f(\bx) \sim \GP(0, \, \kappa(\bx,\bx')), \quad p(\bY \mid \bff) = \prod_{n=1}^{N_t} \prod_{k=1}^{N_p} p(\bY_{n,k} \mid \bff_{n,k}),
	\label{eq:spatio_temporal_gp}
\end{align}
where $\bff_{n,k} = f(\bX_{n,k})$, and we slightly abuse notation by writing $f(\bx) = f(t,\bp)$ and $\kappa(\bx,\bx') = \kappa(t,\bp,t',\bp')$, with $\bp$ denoting the product (or service) features describing their characteristics, including their spatial location when demand is distributed in both space and time, as in the cases of demand for EV charging or Mobility-on-Demand services. Similarly to standard censored regression, our goal is two-fold: i) compute the posterior distribution over the (latent) true demand $\bff$ given the observed demand $\bY$ (censored and subject to substitutions), and ii) predict the true demand $\bff_*$ for a new test point $\bx_*$ (e.g., for forecasting). 

\section{Diffusion-aware Censored Gaussian Processes}

By leveraging the Tobit likelihood in Eq.~\ref{eq:tobit_lik}, Censored GP regression (Section~\ref{subsec:cgp}) allows us to account for the bias induced by censoring by essentially giving the latent function $f$ added flexibility to take higher values for censored observations ($c_n = 1$). However, when modeling aggregate demand data, this approach doesn't account for the exchange of demand that may occur between similar products or services when some of them run out of supply, thus resulting in abnormal demand being observed for ``substitute products'' \citep{wan2018demand,schaafsma2020substitution}. Let $\bff_n = (f_{n,1}, \dots, f_{n,N_p})$ denote a vector containing the true latent aggregate demand for the $N_p$ products or services at time index $n$. We can formally define the concept of unsatisfied demand for a product $k$ as: $\max(f_{n,k} - u_{n,k}, \, 0)$. We propose to model the process of transfer of unsatisfied demand across products as a non-linear diffusion process on a graph. 

\subsection{Transition dynamics}
\label{subsec:transition_dynamics}

Let $\mathcal{G} = (\mathcal{V},\mathcal{E},\bW)$ be a weighted undirected graph, where $\mathcal{V}$ is a set of nodes $|\mathcal{V}| = N_p$ representing products or services, $\mathcal{E}$ is a set of edges, and $\bW \in \mathbb{R}^{N_p \times N_p}$ is weighted adjacency matrix representing the similarity between products. We define the edge weights with the aid of kernel as
\begin{align}
\bW_{i,j} = 
\begin{cases}
\kappa_{\text{diff}}(\bp_i, \, \bp_j) \, , \quad &\text{if} \mbox{ }\mbox{ } i \neq j \\
0 \, , \quad &\text{if} \mbox{ }\mbox{ } i = j
\end{cases}, 
\end{align}
with $\kappa_{\text{diff}}(\bp_i,\bp_j)$ being determined by an RBF kernel:
\begin{align}
	%\kappa_{\text{diff}}(\bp_i, \, \bp_j) = \exp\left\{ - \frac{\|\bp_i-\bp_j\|}{(\ell_{\text{diff}})^2}^2 \right\}.
	\kappa_{\text{diff}}(\bp_i, \, \bp_j) = \exp\left\{ - \|\bp_i-\bp_j\|^2 / \ell_{\text{diff}}^2 \right\}.
\end{align}
The underlying assumption is that products with similar characteristics are likely to be used as replacements when a given product is out of supply, with the lengthscale parameter $\ell_{\text{diff}}$ controlling how much customers are willing to tolerate differences in product characteristics $\bp$. Diffusion of the unsatisfied demand to the same node (self-loops) is obviously not allowed. We then define the state transition matrix as
\begin{align}
	\bT = \text{diag}(\bW \, \textbf{1}_{N_p})^{-1} \, \bW,
	\label{eq:transition_op}
\end{align}
where $\textbf{1}_{N_p}$ denotes an $N_p \times N_p$ matrix of ones. Eq.~\ref{eq:transition_op} essentially produces a row-normalized version of $\bW$, such that the diffusion of unsatisfied demand from service $i$ to service $j \neq i$ becomes: $\bT_{i,j} = \kappa_{\text{diff}}(\bp_i, \, \bp_j) / \sum_{l=1}^{N_p} \kappa_{\text{diff}}(\bp_i, \, \bp_l)$. 
%\begin{align}
%\bT_{i,j} = \frac{\kappa_{\text{diff}}(\bp_i, \, \bp_j)}{\sum_{l=1}^{N_p} \kappa_{\text{diff}}(\bp_i, \, \bp_l)} .
%\end{align}
However, this assumes that all unsatisfied demand at a given node $i$ is transferred to the neighboring nodes $j \in \N_i$, regardless of how different the service characteristics $\bp_i$ are from $\bp_j$. We can relax this assumption by introducing a ``sink'' node in the graph that connects to all the other nodes and extending $\bW$ to be $(N_p+1) \times (N_p+1)$, with $\bW_{i,\text{sink}} = \pi_\text{sink}, \forall i \neq \text{sink}$, while $\bW_{\text{sink},i} = 0$ and $\bW_{\text{sink},\text{sink}} = 1$, in order to ensure that once demand enters the sink node it never leaves. Therefore, the parameter $\pi_\text{sink} \in \mathbb{R}$ controls the likelihood of customers simply giving up and deciding not to consume any alternative product when faced with the lack of supply for their intended product. 

\begin{figure}
    \centering
    \includegraphics[width=.5\textwidth,trim={2.0cm 0.8cm 1.8cm 1cm},clip]{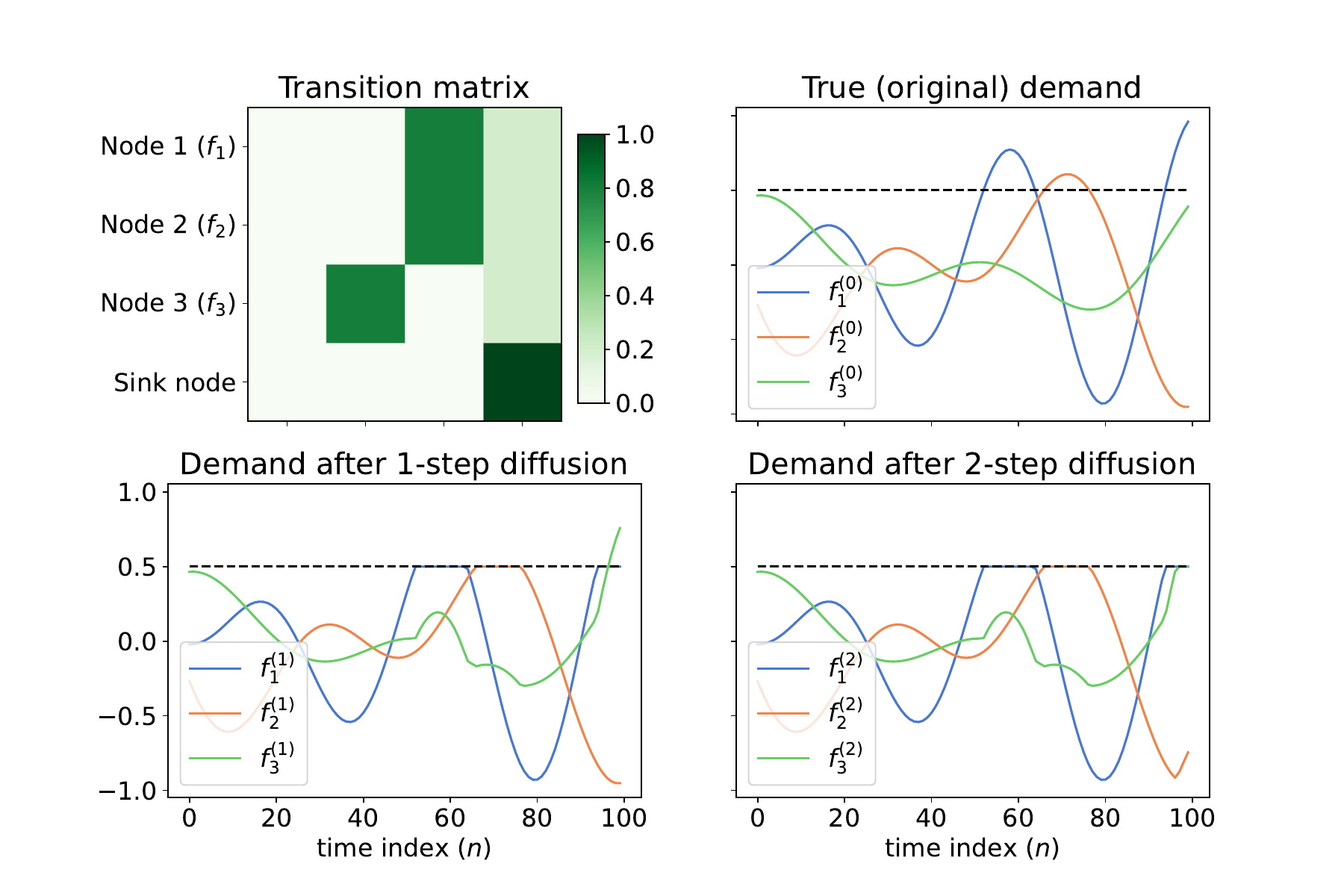}
    \caption{Diffusion process demo for 3 time-series using the transition matrix on the top-left (node 1 sends 80\% of its unsatisfied demand to node 3 and 20\% to the sink node, and so on). The true demand (top-right) that is above the supply (dashed line) gets transferred to the other nodes (incl.~sink node) over the 2 diffusion steps (bottom row).}
    \label{fig:diffusion_demo}
\end{figure}

Based on $\bT$, we propose to model the diffusion dynamics as
\begin{align}
	\bff_n^{(i+1)} &= \bff_n^{(i)} + \underbrace{\max(\bff_n^{(i)} - \bu_n, \, 0) \, \bT}_{\text{incoming demand from others}} - \underbrace{\max(\bff_n^{(i)} - \bu_n, \, 0)}_{\text{outgoing demand}},\nonumber\\
	&= \bff_n^{(i)} + {\max(\bff_n^{(i)} - \bu_n, \, 0)} (\bT-\textbf{I})
	\label{eq:transition_dynamics}
\end{align}
where $\bu_n$ is a vector of the corresponding available supply at time index $n$, thus making $\max(\bff_n^{(i)} - \bu_n, \, 0)$ the unsatisfied demand above the available supply at diffusion step $i$. If we assume that customers have no access to information about the available supply, we can apply the transition operator in Eq.~\ref{eq:transition_dynamics} multiple times in order to simulate the multi-step process of trying to use a service, finding out that there is no available supply, and then searching for an alternative, as can happen, for example, for EV charging and Mobility-on-Demand \citep{unterluggauer2023impact}. In practice, we limit the maximum number of diffusion steps to $N_\text{diff}$. Figure~\ref{fig:diffusion_demo} illustrates two steps of this diffusion process for 3 time-series. 

However, if we assume that customers have access to information about the available supply of all products (incl.~alternatives), then it is unrealistic to consider a multi-step diffusion process. Instead, we propose introducing a time-dependent graph $\mathcal{G}_n = \{\mathcal{V}, \mathcal{E}_n\}$, such that $\mathcal{E}_n = \{ (i,j) \mid u_{n,j} > y_{n,j} \}$, thereby not allowing connections to products that are also out of supply at time index $n$, and applying a single diffusion step (Eq.~\ref{eq:transition_dynamics}) in $\mathcal{G}_n$. This modeling approach encodes the intuitive assumption that customers, when faced with a product out of supply, will look at all the available alternatives once and choose one or none (sink node). 

\subsection{Likelihood}

After applying the graph diffusion process $N_\text{diff}$ times, the additional demand observed at time index $n$ due to the demand transfer process caused by the limited supply of other products or services can be computed as
\begin{align}
	%\bd_n = \bff_n + \underbrace{\max(\bff_n^{(N_\text{diff})} - \bff_n, 0)}_{\text{addtional demand due to substitutions}}. 
	\bd_n = \max(\bff_n^{(N_\text{diff})} - \bff_n, 0). 
\end{align}
Based on this ``diffused demand'' $\bd_n$, we propose to model the likelihood of the observed demand values, accounting for censoring and for the diffusion of unsatisfied demand, as
\begin{align}
	%p(\bY_n | \bff_n) &= \prod_{k=1}^{N_p} \N(y_{n,k} | f_{n,k} + d_{n,k}, \sigma^2)^{\mathbbm{1}(y_{n,k} < u_{n,k})} \nonumber\\
	%&\times (1-\Phi(y_{n,k} | f_{n,k} + d_{n,k}, \sigma^2))^{\mathbbm{1}(y_{n,k} \geq u_{n,k})}, 
	p(\bY_n \mid \bff_n) &= \prod_{k=1}^{N_p} \N(y_{n,k} \mid f_{n,k} + d_{n,k}, \sigma^2)^{(1-c_{n,k})} \nonumber\\
	&\quad \times (1-\Phi(y_{n,k} \mid f_{n,k} + d_{n,k}, \sigma^2))^{c_{n,k}}. 
	\label{eq:diffused_lik}
\end{align}
%where $\mathbbm{1}(\cdot)$ is an indicator function that takes the value $1$ when the condition is true, and zero otherwise. 
This likelihood can be intuitively understood as a combination of: i) a Tobit formulation to account for the observed demand for an out-of-supply product $k$ ($c_{n,k} = 1$) potentially being lower than the latent process expected it to be, and ii) an added term $d_{n,k}$ to the expected demand value of product $k$ to account for the potentially extra (unexpected) demand observed due to similar products being out of supply. Kindly note that although Eq.~\ref{eq:diffused_lik} factorizes across services $k$, the likelihoods for different products are no longer independent due to the diffusion process, which ties them together. Also, note that using the likelihood in Eq.~\ref{eq:diffused_lik} implies access to information about the available supply $\bu_n$, which is required for the diffusion process (Eq.~\ref{eq:transition_dynamics}). This contrasts with regular Tobit regression (Eq.~\ref{eq:tobit_lik}), where only access to a censoring indicator variable $\bc_n$ is required. Nevertheless, we argue that this is a reasonable assumption since, for most real-world demand modeling applications, supply information is typically readily available along with the demand data. Lastly, it should be noted that although we use Gaussian processes as the backbone for our proposed approach, the methodology described above can be generalized to other modeling approaches such as neural networks and linear models. 

\subsection{Inference}

Scalability is often a concern when modeling large datasets with GPs. In our case, we are interested in jointly modeling potentially-long time-series of observed demand across different products or services, thus further aggravating this concern, since computing the posterior distribution $p(\bff \mid \bX)$ typically has a cubic cost of $\mathcal{O}(N_t^3 N_p^3)$. Therefore, we leverage the work of \citet{hamelijnck2021spatio}, which combines spatio-temporal filtering with natural gradient variational inference to achieve linear scalability with respect to time. %Although the authors focus on spatiotemporal data, we adapt their proposed framework to model aggregate demand data for various related services or products over time. 

We begin by assuming that our kernel is both Markovian and separable between time and products, i.e.~$\kappa(t,\bp,t',\bp') = \kappa_t(t,t') \, \kappa_\bp(\bp,\bp')$. Under this assumption, the model in Eq.~\ref{eq:spatio_temporal_gp} can be reformulated as a state space model, thus reducing the computational scaling to linear in $N_t$. The GP prior can be written as a stochastic differential equation (SDE) \citep{chang2020fast}, which, in this case, requires marginalizing to a finite set of products, $\textbf{P} \in \mathbb{R}^{N_p \times D_p}$, giving, $\text{d} \bar{\bff}(t) = \textbf{F} \, \bar{\bff}(t) \, \text{d}t + \textbf{L} \, \text{d}\bs\beta(t)$, where $\bar{\bff}(t)$ is the Gaussian distributed state over products $\textbf{P}$ at time $t$, %, and with dimensionality $d = N_p d_t$, where $d_t$ is the dimensionality of the state space induced by the temporal kernel $\kappa_t(t,t')$. 
and $\textbf{F}$ and $\textbf{L}$ are the feedback and noise effect matrices. The function value $\bff_n$ can be extracted from the state by a matrix $\textbf{H}$ as $\bff_n = \textbf{H} \, \bar{\bff}(t_n)$. For a fixed step size $\Delta_n = t_{n+1} - t_n$, the resulting discrete model is \citep{chang2020fast}:
\begin{align}
\bar{\bff}(t_{n+1}) &= \textbf{A}_n \bar{\bff}(t_{n}) + \textbf{q}_n, \nonumber\\ 
\bY_n \mid \bar{\bff}(t_n) &\sim p(\bY_n \mid \textbf{H} \, \bar{\bff}(t_n)),
\label{eq:linear_model}
\end{align}
where $\textbf{A}_n = \exp(\textbf{F} \Delta_n)$ is the linear state transition matrix, and $\textbf{q}_n \sim \mathcal{N}(\textbf{0}, \textbf{Q}_n)$, with $\textbf{Q}_n$ denoting the process noise covariance. 
%and the model matrices are given by
%\begin{align}
%\textbf{A}_n &= \bs\Psi(\textbf{F} \Delta_n), \\ 
%\textbf{Q}_n &= \int_0^{\Delta_n} \bs\Psi(\Delta_n - \tau) \, \textbf{L} \, \textbf{Q}_c \, \textbf{L}^\top \, \bs\Psi(\Delta_n - \tau)^\top \, \text{d}\tau,
%\end{align}
%with $\bs\Psi(\cdot)$ denoting the matrix exponential. 
If $p(\bY_n \mid \textbf{H} \, \bar{\bff}(t_n))$ is Gaussian, then Kalman smoothing algorithms to be employed to perform inference in linear time in $N_t$. Since the likelihood of our proposed model (Eq.~\ref{eq:diffused_lik}) is non-Gaussian, we apply conjugate-computation variational inference (CVI) \citep{khan2017conjugate}, whereby an \emph{approximate likelihood}, that is conjugate to the prior (in this particular case, Gaussian), with free parameters $\tilde{\bs\lambda}$, is considered to enable efficient computation of the natural gradients of the evidence lower bound (ELBO). 

As shown by \citet{khan2017conjugate}, if the chosen approximate likelihood is conjugate to the prior, then performing natural gradient VI is equivalent to a two-step Bayesian update of the form:
\begin{align}
	\tilde{\bs\lambda} \leftarrow (1-\beta) \tilde{\bs\lambda} + \beta \frac{\partial \E_{q(\bff)}[\log \mbox{ } p(\bY \mid \bff)]}{\partial \bs\mu},  \quad\quad \bs\lambda \leftarrow \bs\eta + \tilde{\bs\lambda},
	\label{eq:cvi_updates}
\end{align}
where $\bs\lambda$ are the natural parameters of the approximate posterior distribution (with corresponding mean parameters $\bs\mu$), $\bs\eta$ are the natural parameters of the prior, and $\tilde{\bs\lambda}$ are the natural parameters of the (approximate) likelihood contributions. 

Let $\N(\tY \mid \bff, \tV)$ be an approximate likelihood parameterised by covariance $\tV = (-2\tilde{\bs\lambda}^{(2)})^{-1}$ and mean $\tY = \tV \tilde{\bs\lambda}^{(1)}$, with $\tilde{\bs\lambda} = \{\tilde{\bs\lambda}^{(1)}, \tilde{\bs\lambda}^{(2)}\}$. The approximate posterior then has the form $q(\bff) \propto \N(\tY \mid \bff, \tV) \, p(\bff)$. Since the GP prior is Markov and the approximate likelihood factorizes across time \citep{hamelijnck2021spatio}, the approximate GP posterior is also Markov \citep{tebbutt2021combining}. Therefore, and since the approximate likelihood is now Gaussian, the marginals $q(\bff_n)$ can be computed through linear filtering and smoothing applied to Eq.~\ref{eq:linear_model}, but with $\tY_n \mid \bar{\bff}(t_n) \sim \N(\tY_n \mid \textbf{H} \, \bar{\bff}(t_n), \tV)$ as the measurement model. Given the marginals, Eq.~\ref{eq:cvi_updates} can be used to give the new likelihood parameters $\tY$ and $\tV$. The ELBO can be written as a sum of three terms:
\begin{align}
	\mathcal{L} &= \E_{q(\bff)}\left[ \log \mbox{ } \frac{p(\bY \mid \bff) \, p(\bff)}{q(\bff)}\right] = \sum_{n=1}^{N_t} \E_{q(\bff_n)}[\log \mbox{ } p(\bY_n \mid \bff_n)]\nonumber\\  &\quad- \E_{q(\bff)}[\log \mbox{ } \N(\tY \mid \bff, \tV)] + \log \mbox{ } \E_{p(\bff)}[\N[\tY \mid \bff, \tV]], 
	\label{eq:elbo}
\end{align}
where the first term corresponds to the expected log likelihood, the second term is the expected log \emph{approximate likelihood}, and the third is the log marginal likelihood of the approximation posterior. As shown in \citep{hamelijnck2021spatio}, these terms can also be computed efficiently based on the outputs of the filtering and smoothing algorithms. Computing the updates to the variational parameters in Eq.~\ref{eq:cvi_updates}, as well as evaluating the ELBO (Eq.~\ref{eq:elbo}), requires computing $\E_{q(\bff_n)}[\log p(\bY_n \mid \bff_n)]$ and its gradients, for which we rely on quadrature. The gradients are back-propagated through the graph diffusion process using automatic differentiation. 

\subsection{Diffusion parameters learning}
\label{subsec:diff_param_learning}

Along with the hyper-parameters from the kernels $\kappa_t$ and $\kappa_{\bp}$ of the GP prior, the proposed modeling approach introduces additional hyper-parameters to the likelihood due to the embedded diffusion process. The hyper-parameters of the likelihood in Eq.~\ref{eq:diffused_lik} are then: the observation variance $\sigma^2$, the diffusion lengthscale $\ell_\text{diff}$, the sink node parameter $\pi_\text{diff}$, and the number of diffusion steps $N_\text{diff}$. While we assume $N_\text{diff}$ to be fixed, the former three can also be learned from data. Although, for most domains, there probably is domain knowledge to guide the choice of the diffusion lengthscale $\ell_\text{diff}$ and the sink node parameter $\pi_\text{diff}$ due to their natural interpretations from a behavioral point of view, in some scenarios, a data-driven approach may be preferred. In practice, this can be achieved by, for example, setting these hyper-parameters to reasonable initial values, and then performing type-II maximum likelihood estimation via maximizing the ELBO in Eq.~\ref{eq:elbo} with gradient descent updates alternated with the natural gradient updates in Eq.~\ref{eq:cvi_updates}. 

\section{Experiments}

In this section, we empirically demonstrate the capabilities of the proposed Diffusion-aware Censored GP (``DCGP''), focusing on its ability to infer the true demand based on demand observations that are censored and subject to substitutions, as well as its ability to produce more accurate predictions of true demand. We consider two main variants of the proposed approach - one with fixed $\ell_{\text{diff}}$ and another where $\ell_{\text{diff}}$ is learned (referred to as ``DCGP-f'' and ``DCGP-l'', respectively), which we compare with Censored GPs (``CGP'') and standard (Non-Censored) GPs (``NCGP'') fitted to the observed demand. For reference, we also provide results for an ``Oracle'' GP fitted to the true (uncensored) demand. We compare the different modeling approaches across three metrics: RMSE, R-squared ($\text{R}^2$), and negative log predictive density (NLPD). Source code for all the experiments is provided at: \url{https://github.com/fmpr/Diffusion-aware-Censored-GP}

\subsection{Artificial data}
\label{subsection:artificial_data}

\begin{figure*}[!ht]
    \centering
    \subfigure[Data generation]{\includegraphics[width=0.49\textwidth,trim={2.8cm 0cm 3cm 1.2cm},clip]{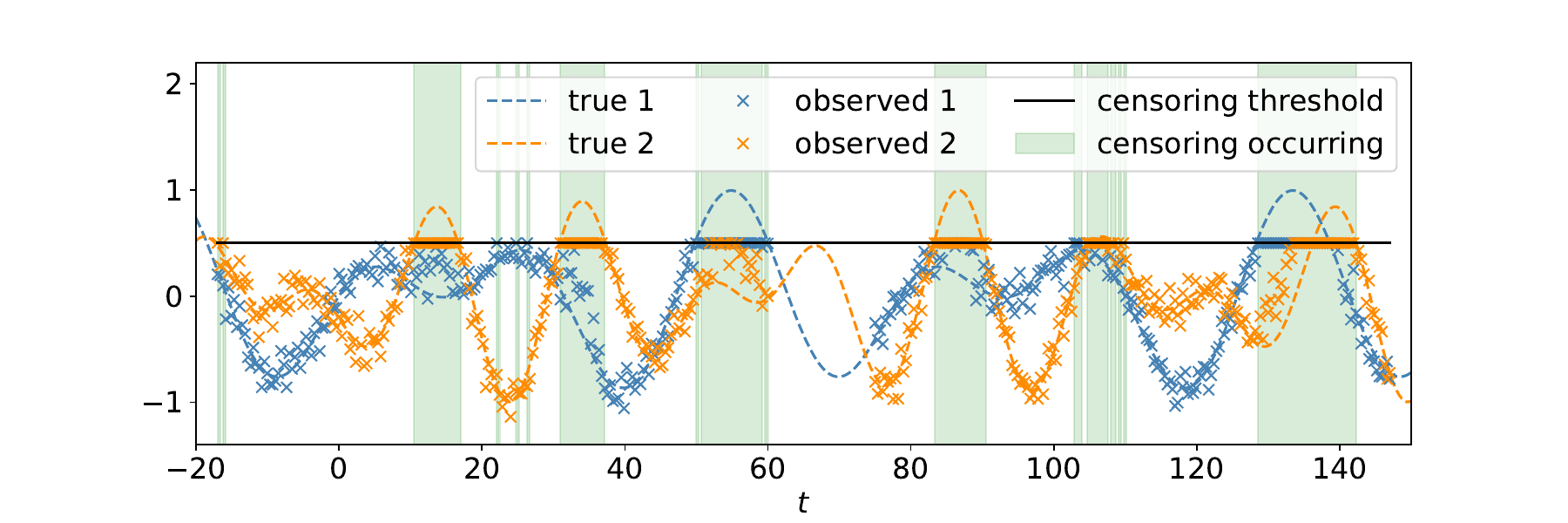}} 
    \subfigure[(Non-Censored) GP posterior]{\includegraphics[width=0.49\textwidth,trim={2.8cm 0cm 3cm 1.2cm},clip]{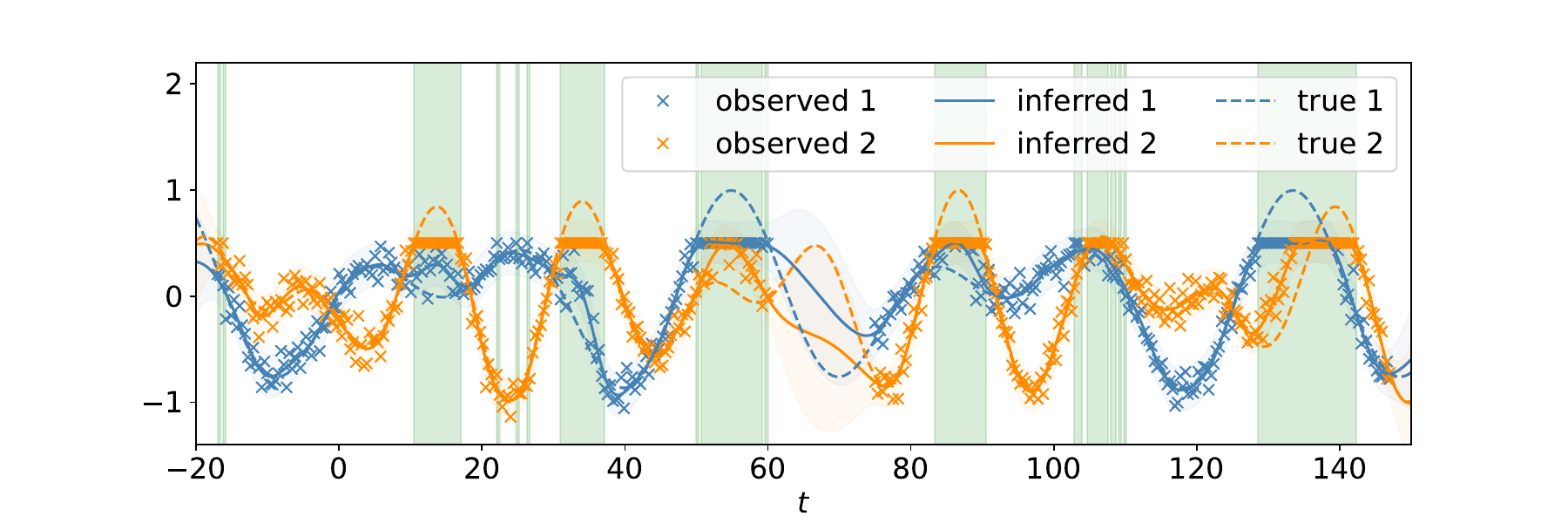}} \\\vspace{-0.2cm}
    \subfigure[Censored GP posterior]{\includegraphics[width=0.49\textwidth,trim={2.8cm 0cm 3cm 1.2cm},clip]{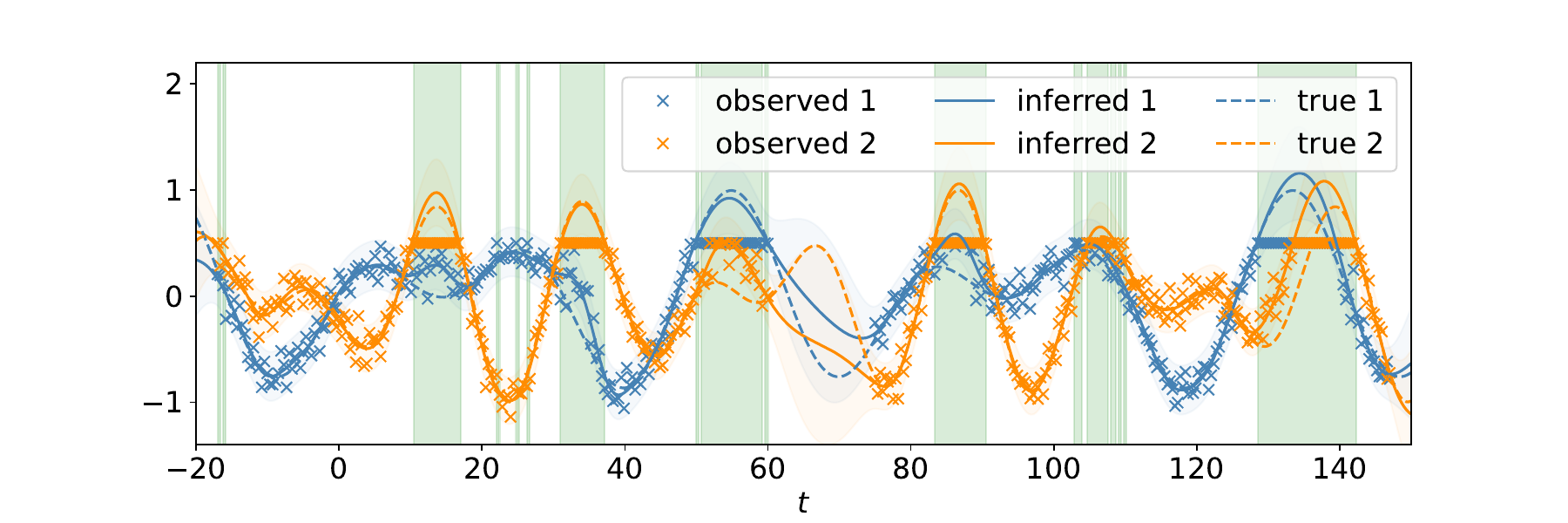}}
    \subfigure[Diffused Censored GP posterior]{\includegraphics[width=0.49\textwidth,trim={2.8cm 0cm 3cm 1.2cm},clip]{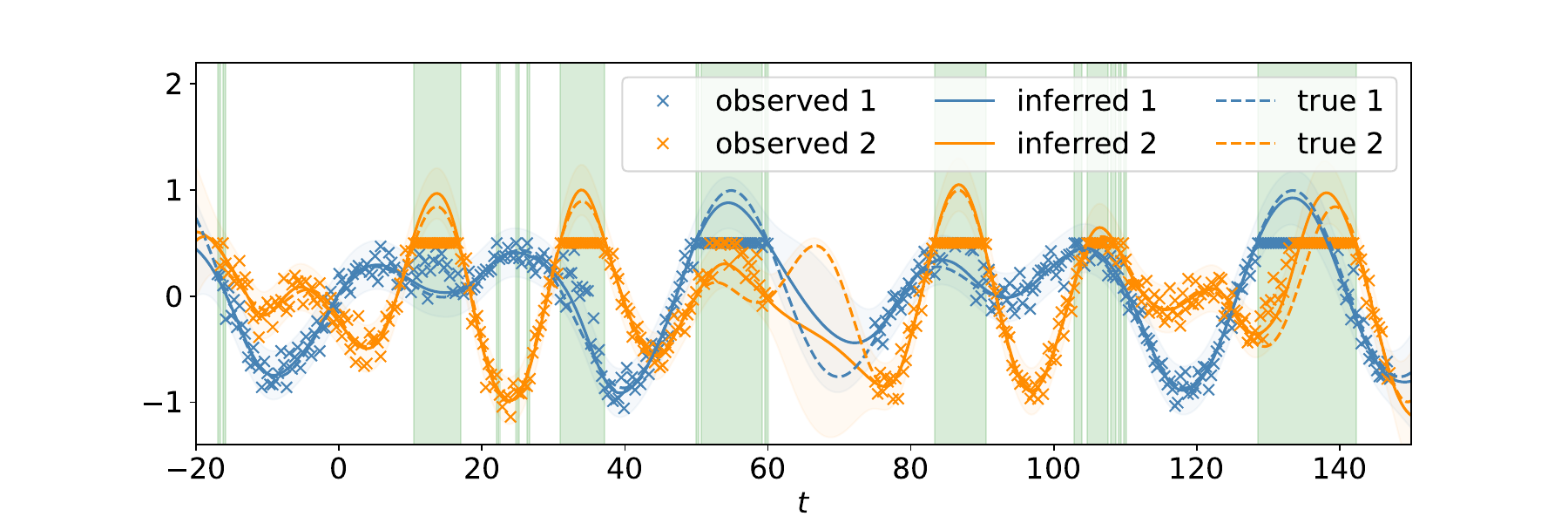}}\vspace{-0.3cm}
    \caption{GP posteriors obtained by the different models for the artificial dataset (Dataset A) depicted in (a). The green-shaded areas indicate time indexes where at least one of the time-series is subject to censoring. As shown in (d), the proposed approach achieves the best approximation to the true demand.}
    \label{fig:fake_indep_data_results}
\end{figure*}

\textbf{Indepedent time-series.} We begin by demonstrating the proposed approach using independently generated time-series data based on sinusoidal functions. We construct three datasets. In Dataset A, we draw samples from two sinusoidal functions (simulating the true demand) with some observation (Gaussian) noise added, and then simulate the censoring process by setting a constant censoring threshold (supply). The demand above the threshold is sent to the other time-series, and if the resulting demand for the other time-series now becomes higher than the threshold, then we throw away the unsatisfied demand (see Figure~\ref{fig:fake_indep_data_results}a). Dataset B is similar to Dataset A, but the threshold is sampled randomly from a uniform distribution for each time step, thus resulting in a stochastic supply that causes very sporadic supply shortages rather than shortages lasting for longer time intervals as in Dataset A. Dataset C is similar to Dataset A but considers three sinusoidal time-series (instead of two) mimicking the demand of three products. Whenever Product 1 or Product 3 are out of supply, their unsatisfied demand goes to Product 2, while the unsatisfied demand from the latter is distributed equally among Products 1 and 3. All models use the same underlying GP framework with independent GPs for each product (i.e.~$\kappa_\bp(\bp,\bp') = \mathbbm{1}(\bp = \bp')$) and $\kappa_t(t,t')$ being a Matern kernel. Details are in Appendix~\ref{appendix:fake_indep_data}.  

Table~\ref{tab:fakeIndepData_indepKernel} shows the obtained test set results (a more complete depiction of the results is provided in Appendix~\ref{appendix:analysis_inferred_demand}). As expected, the Non-Censored GP performs the worst across the three datasets by simply fitting the observed data (see, e.g.,  Figure~\ref{fig:fake_indep_data_results}b). The Censored GP is able to infer that the demand should be higher than observed at censored locations (see Figure~\ref{fig:fake_indep_data_results}c), thus achieving better results when compared to the Non-Censored GP in Table~\ref{tab:fakeIndepData_indepKernel}. However, as Figure~\ref{fig:fake_indep_data_results}c shows, the Censored GP is unable to understand that the abnormally high demand observed in some time-series can be explained by the transfer of unsatisfied demand from other time-series that are subject to censoring. On the other hand, the proposed approach is able to account for this transfer and, to a great extent, recover the true underlying demand (Figure~\ref{fig:fake_indep_data_results}d), which can be also observed in the results in Table~\ref{tab:fakeIndepData_indepKernel}. Interestingly, for Datasets A and C, the best results were obtained with a fixed $\ell_{\text{diff}} = 1$, while for Dataset B learning $\ell_{\text{diff}}$ resulted in slightly better results. %, which we attribute to the fact that the sporadic supply shortages present in this dataset constitute a simpler problem, while in Datasets A and C the supply shortages occur in longer bursts, thus causing the underlying GP model to be less confident in ``explaining'' them through the diffusion process. 
Across all datasets, the proposed approach is able to achieve results close to the ``True GP (Oracle)'' reference, thus demonstrating its ability to infer the true underlying functions from the observations. 

\begin{table}[!t]
    \centering
    \small
    \begin{tabular}{p{0.1cm}l | p{0.73cm} p{0.73cm} p{0.73cm} p{0.73cm}}
    %\begin{tabular}{cl | rrrr}
        \toprule
	& & & & & $\textbf{RMSE}$\\
	&\textbf{Model} & \textbf{NLPD} & \textbf{RMSE} & $\textbf{R}^2$ & $\textbf{funct.}$ \\ 
        \midrule
         \multirow{5}{*}{\rotatebox[origin=c]{90}{Dataset A}} & True GP (Oracle) & -1.771 & 0.099 & 0.960 & 0.029\\
         & NCGP & 0.853 & 0.188 & 0.858 & 0.163\\
         & CGP & -0.674 & 0.162 & 0.895 & 0.130\\
         & DCGP-f & {-0.685} & \textbf{0.108} & \textbf{0.953} & \textbf{0.051}\\
         & DCGP-l & \textbf{-0.751} & {0.118} & {0.944} & {0.071}\\
        \midrule
         \multirow{5}{*}{\rotatebox[origin=c]{90}{Dataset B}} & True GP (Oracle) & -1.701 & 0.103 & 0.957 & 0.029\\
         & NCGP & -0.505 & 0.188 & 0.857 & 0.158\\
         & CGP & \textbf{-1.004} & 0.148 & 0.912 & 0.111\\
         & DCGP-f & 0.911 & {0.107} & \textbf{0.954} & {0.038}\\
         & DCGP-l & {0.590} & \textbf{0.106} & \textbf{0.954} & \textbf{0.037}\\
        \midrule
         \multirow{5}{*}{\rotatebox[origin=c]{90}{Dataset C}} & True GP (Oracle) & -2.527 & 0.104 & 0.954 & 0.031\\
         & NCGP & 0.805 & 0.188 & 0.856 & 0.162\\
         & CGP & -1.224 & 0.157 & 0.900 & 0.122\\
         & DCGP-f & -1.311 & \textbf{0.107} & \textbf{0.952} & \textbf{0.042}\\
         & DCGP-l & \textbf{-1.380} & {0.130} & {0.930} & {0.086}\\
        \bottomrule
    \end{tabular}
    \caption{Experimental results obtained for the three artificial datasets of independently-generated (uncorrelated) time-series. ``RMSE funct.'' further measures the RMSE to the true underlying (noiseless) function.} %, thus measuring the ability to recover the true function.} 
    %Inclides: paper - demo\_fakeIndepData1\_indepKernel\_throwExcess.ipynb, paper - demo\_fakeIndepData2\_indepKernel\_throwExcess.ipynb, paper - demo\_fakeIndepData3\_indepKernel\_throwExcess.ipynb}
    \label{tab:fakeIndepData_indepKernel}
\end{table}

\textbf{Spatio-temporal data.} We now consider artificial spatio-temporal demand data, whereby demand varies across space and time as, for example, in the cases of EV charging demand or Mobility-on-Demand services. We generate artificial true demand data by first sampling $N_p=10$ 2-D spatial locations uniformly in the range $[-2,2]$, and then sampling 10 time-series of length $N_t=400$ from a spatio-temporal GP with a separable kernel with a Periodic+Matern component over time and a Matern component over space, to which we then add Gaussian noise. We simulate the censoring process using a 2-state Markov model, where the states represent ``with supply'' and ``out of supply'' states, with the latter employing a fixed demand threshold. The unsatisfied demand is transferred to the nearby locations by simulating the diffusion process described in Section~\ref{subsec:transition_dynamics} with fixed diffusion parameters and with a sink node probability of $20\%$. In this scenario, the product features $\bp$ correspond to the spatial locations. We generate two datasets: ``Dataset ST-1'' and ``Dataset ST-2'' using $N_{\text{diff}}=1$ and $N_{\text{diff}}=2$ diffusion steps, respectively. We use 130 evenly-sampled observations for training and the remaining for testing. We provide additional details, including figures, in Appendix~\ref{appendix:fake_ST_data}. 

\begin{table}[!t]
    \centering
    \small
    \begin{tabular}{p{0.1cm}l | p{0.73cm} p{0.73cm} p{0.73cm} p{0.73cm}}
    %\begin{tabular}{cl | rrrr}
        \toprule
	& & & & & $\textbf{RMSE}$\\
	&\textbf{Model} & \textbf{NLPD} & \textbf{RMSE} & $\textbf{R}^2$ & $\textbf{funct.}$ \\ 
        \midrule
         \multirow{5}{*}{\rotatebox[origin=c]{90}{Dataset ST-1}} & True GP (Oracle) & -0.031 & 0.273 & 0.924 & 0.178\\
         & NCGP & 0.327 & 0.368 & 0.869 & 0.313\\
         & CGP & 0.111 & 0.313 & 0.900 & 0.239\\
         & DCGP-f & \textbf{-0.019} & \textbf{0.268} & \textbf{0.926} & \textbf{0.172}\\
         & DCGP-l & {0.047} & {0.281} & {0.920} & {0.191}\\
        \midrule
         \multirow{7}{*}{\rotatebox[origin=c]{90}{Dataset ST-2}} & True GP (Oracle) & -0.031 & 0.273 & 0.924 & 0.178\\
         & NCGP & 0.285 & 0.358 & 0.877 & 0.300\\
         & CGP & 0.140 & 0.318 & 0.899 & 0.246\\
         & DCGP-f (1-step) & {0.020} & 0.282 & 0.920 & 0.193\\
         & DCGP-l (1-step) & {0.088} & {0.302} & {0.909} & {0.223}\\
         & DCGP-f (2-step) & \textbf{0.018} & \textbf{0.269} & \textbf{0.926} & \textbf{0.172}\\
         & DCGP-l (2-step) & 0.060 & 0.286 & 0.918 & 0.199\\
        \bottomrule
    \end{tabular}
    \caption{Experimental results obtained for the two artificially-generated spatio-temporal datasets. ``RMSE funct.'' further measures the RMSE to the true underlying (noiseless) function.}
    %Includes: paper - demo\_fakeSTdata9\_STkernel.ipynb, paper - demo\_fakeSTdata10\_STkernel.ipynb}
    \label{tab:fakeSTdata_STkernel}
\end{table}

The results obtained are summarized in Table~\ref{tab:fakeSTdata_STkernel} (see Appendix~\ref{appendix:analysis_inferred_demand} for additional details). Despite the higher dimensionality of the problem and the correlations between time-series, the results remain similar to the smaller independent time-series experiment above. The Censored GP outperforms its Non-Censored counterpart, while the proposed approach further outperforms both (we plot examples of the inferred true demand by the different approaches in Appendix~\ref{appendix:analysis_inferred_demand}). Importantly, even in the more challenging setting where $\ell_{\text{diff}}$ has to be learned, the proposed approach is still able to clearly outperform the other baselines. The non-learnable variant (fixed at $\ell_{\text{diff}} = 0.1$) is able to further outperform it, which is unsurprising given that the data was also generated with $\ell_{\text{diff}} = 0.1$. Therefore, this version corresponds to the scenario where domain knowledge can accurately drive the choice of $\ell_{\text{diff}}$. We provide an in-depth study of the sensitivity to model misspecification in Appendix~\ref{appendix:model_misspec}. In Appendix~\ref{appendix:learning_pi}, we demonstrate the ability to also learn the hyper-parameter $\pi_{\text{diff}}$. Furthermore, the results in Table~\ref{tab:fakeSTdata_STkernel} also demonstrate that the proposed approach is able to effectively take into account multiple steps of diffusion, thus being able to further improve the true demand estimation in Dataset ST-2.

\subsection{Supermaket sales}
\label{subsec:sales_main}

We now turn to experiments on real-world supermarket vegetable sales data obtained from \citep{SupermarketKaggle}. Unfortunately, it is impossible to have access to the true demand, as that would imply having access to the customers' intentions. Therefore, as it is standard practice in the censored demand modeling \citep{gammelli2020estimating,huttel2022modeling} and, more broadly, in the censored regression literature \citep{schmee1979simple,li2003empirical}, we resort to assuming the observed demand to correspond to the true demand, and simulate the censoring process based on it in a realistic manner. Concretely, we consider the demand time-series for two products and simulate supply shortages using a 2-state Markov model similar to the one used for the artificial spatio-temporal data. The unsatisfied demand for one product is then transferred to the other product. If the other product is also out of supply, then that demand is lost. We randomly sample 90\% observations for training and leave 10\% for testing. Additional experimental setup details are provided in Appendix~\ref{appendix:sales_details}. We also experimented with different splits to study the impact of train set size - see Appendix~\ref{appendix:sales_splits}.

\begin{table}[!h]
    \centering
    \small
    \begin{tabular}{l | p{0.73cm} p{0.73cm} p{0.73cm} | p{0.73cm} p{0.73cm} p{0.73cm}}
        \toprule
	& \multicolumn{3}{c}{\textbf{Trainset}} & \multicolumn{3}{|c}{\textbf{Testset}} \\
         \textbf{Model} & \textbf{NLPD} & \textbf{RMSE}  & $\textbf{R}^2$ & \textbf{NLPD} & \textbf{RMSE} & $\textbf{R}^2$ \\ 
%        \midrule
%         True GP (Oracle) & 1.541 & 0.527 & 0.726 & 1.692 & 0.551 & 0.627\\
%         GP & 2.306 & 0.739 & 0.464 & 2.126 & 0.675 & 0.436\\
%         Censored GP & 2.205 & 0.703 & 0.514 & 2.076 & 0.647 & 0.482\\
%         Diffused Censored GP (fixed) & \textbf{1.755} & 0.602 & 0.643 & \textbf{1.962} & 0.607 & 0.545\\
%         Diffused Censored GP (learned) & {2.011} & \textbf{0.592} & \textbf{0.656} & {2.159} & \textbf{0.588} & \textbf{0.574}\\
        \midrule
         True GP & 1.541 & 0.527 & 0.726 & 1.692 & 0.551 & 0.627\\
         NCGP & 2.285 & 0.735 & 0.469 & 2.066 & 0.669 & 0.446\\
         CGP & 2.166 & 0.714 & 0.499 & 2.010 & 0.653 & 0.472\\
         %Diffused Censored GP (fixed) & \textbf{1.748} & 0.591 & 0.657 & \textbf{1.960} & 0.601 & 0.554\\
         DCGP-f & \textbf{1.740} & 0.591 & 0.657 & \textbf{1.968} & 0.602 & 0.553\\
         %Diffused Censored GP (learned) & {2.021} & \textbf{0.620} & \textbf{0.622} & {2.112} & \textbf{0.606} & \textbf{0.546}\\
         DCGP-l & {1.896} & \textbf{0.612} & \textbf{0.632} & {2.103} & \textbf{0.601} & \textbf{0.554}\\
        \bottomrule
    \end{tabular}
    \caption{Experimental results obtained for supermarket sales dataset.}
    %paper - demo\_salesData9010\_indepKernel\_throwExcess.ipynb}
    \label{tab:sales9010}
\end{table}

Table~\ref{tab:sales9010} shows the obtained results for the train and test sets. Note that, in censored regression, train set performance is particularly important, as it is an indicator of the ability to infer the true demand from the observed data. As the results in Table~\ref{tab:sales9010} show, the proposed approach is able to infer the true demand more accurately than the other baselines, thus resulting in substantially better train and test set performance. The additional results with different train/test splits in Appendix~\ref{appendix:sales_splits} further show that the superior performance of the proposed approach is observable even for smaller train sets, where some signs of overfitting start to emerge. 

\subsection{Bike sharing demand}
\label{subsec:bikesharing_main}

We now consider a real-world problem where demand varies across space and time. Concretely, we consider the problem of building a model of the true demand for a hub-based bike-sharing system, where users rent bicycles on an on-demand basis from one of the many hubs available across the city through the use of an app. Bike-sharing systems are known to suffer from strong demand-supply imbalances (e.g., due to normal commuting patterns), thus leading to hubs frequently running out of bicycles. Naturally, if a hub is out of bicycles, the user either walks to the nearest hub with available bicycles or chooses an alternative transportation mode. Therefore, the problem of inferring and predicting the true demand is pivotal for daily operations (e.g., rebalancing \citep{liu2016rebalancing}) and for long-term strategic decisions (e.g., capacity planning and service expansion \citep{liu2017functional}).

\begin{table}[!t]
    \centering
    \small
    \begin{tabular}{l | p{0.73cm} p{0.73cm} p{0.73cm} | p{0.73cm} p{0.73cm} p{0.73cm}}
        \toprule
	& \multicolumn{3}{c}{\textbf{Trainset}} & \multicolumn{3}{|c}{\textbf{Testset}} \\
         \textbf{Model} & \textbf{NLPD} & \textbf{RMSE}  & $\textbf{R}^2$ & \textbf{NLPD} & \textbf{RMSE} & $\textbf{R}^2$ \\ 
        \midrule
         True GP & 2.091 & 1.220 & 0.984 & 3.208 & 6.048 & 0.436\\
         NCGP & 2.674 & 4.016 & 0.805 & 3.297 & 6.603 & 0.332\\
         CGP & 2.552 & 4.197 & 0.769 & 3.252 & 6.436 & 0.365\\
         DCGP-f & 2.328 & \textbf{2.983} & \textbf{0.885} & \textbf{3.171} & \textbf{5.977} & 0.448\\
         DCGP-l & \textbf{2.311} & {2.992} & {0.882} & 3.175 & 5.982 & \textbf{0.449}\\
        \bottomrule
    \end{tabular}
    \caption{Experimental results obtained for bike-sharing dataset.}
    %paper - demo\_bikeData7030\_STkernel\_4ST\_v3\_throwExcess.ipynb}
    \label{tab:bike_4stations}
\end{table}

Since no ground truth data is available for this problem, we follow the same approach as in \citep{gammelli2020estimating} to simulate the censoring process based on the supply data and the observed demand data for 4 bicycle hubs in NYC - which is assumed to correspond to the true demand. The unsatisfied demand above the available supply is transferred to other hubs using the diffusion process from Section~\ref{subsec:transition_dynamics}. Details are provided in Appendix~\ref{appendix:bike_details}. The obtained results in Table~\ref{tab:bike_4stations} again clearly show that the proposed approach is able to more accurately infer the true demand (as indicated by the train set performance), even in the more challenging case where $\ell_{\text{diff}}$ is unknown. The improved estimate of the true demand for the train set then translates into more accurate predictions of the true demand in the test set. Interestingly, in this experiment, the Censored GP performs worse than its Non-Censored counterpart in the train set, which could be explained by its inability to account for the spillover effect to nearby hubs whenever a hub is out of supply. We also consider a simpler version of this experiment with just 2 hubs, which is easier to analyze, in Appendix~\ref{appendix:bike_2stations}.

\subsection{EV charging demand}

Lastly, we consider the problem of inferring the true spatio-temporal demand for EV charging from observations consisting of (realized) charging events. We use the agent-based simulation tool, GAIA \citep{unterluggauer2023impact}, to generate EV charging data for the whole commune of Frederiksberg in Copenhagen. GAIA is based on the notion of a steady-state SoC distribution \citep{hipolito2022charging} and a probabilistic decision-to-charge model to be able to simulate and analyze different charging strategies, thereby addressing the uncertainties resulting from the additional demand for EV charging by accounting for home charging availability, charging location, and the decision to charge in space and time. Since GAIA was carefully calibrated with real-world data sources, such as data from the energy distribution network and from the Danish National Travel Survey, and because it models individual agent behavior, it provides a unique environment for validating our proposed approach by allowing access to the true demand along with the realized (observable) demand. We use GAIA to simulate 4 weeks of EV charging event data. We partition the area of Frederiksberg into 9 clusters using k-means and time in 5-minute bins. We hold out 30\% of data for testing. Appendix~\ref{appendix:EV_charging_details} provides additional details on the experimental setup. 

\begin{table}[!t]
    \centering
    \small
    \begin{tabular}{l | p{0.73cm} p{0.73cm} p{0.73cm} | p{0.73cm} p{0.73cm} p{0.73cm}}
        \toprule
	& \multicolumn{3}{c}{\textbf{Trainset}} & \multicolumn{3}{|c}{\textbf{Testset}} \\
         \textbf{Model} & \textbf{NLPD} & \textbf{RMSE}  & $\textbf{R}^2$ & \textbf{NLPD} & \textbf{RMSE} & $\textbf{R}^2$ \\ 
        \midrule
         True GP & 1.059 & 0.339 & 0.965 & 1.111 & 0.443 & 0.942\\
         NCGP & 1.806 & 1.408 & 0.407 & 1.817 & 1.424 & 0.397\\
         CGP & 1.721 & 1.357 & 0.447 & 1.740 & 1.375 & 0.431\\
         DCGP-f & \textbf{1.599} & \textbf{1.215} & \textbf{0.562} & \textbf{1.628} & \textbf{1.244} & \textbf{0.540}\\ % lengthscale=100, diff variance=4, lr_adam = 1.0, lr_newton = 0.1 - BUT WITH periodic lengthscale = 0.1
         DCGP-l & \textbf{1.611} & \textbf{1.222} & \textbf{0.558} & \textbf{1.639} & \textbf{1.251} & \textbf{0.536}\\
        \bottomrule
    \end{tabular}
    \caption{Experimental results obtained for EV charging dataset.}
    %paper - demo\_GAIA\_STkernel\_4weeks-aggregated\_ts copy 7.ipynb}
    \label{tab:GAIA}
\vspace{-0.3cm}
\end{table}

Table~\ref{tab:GAIA} shows the obtained results. Although the Censored GP already shows some improvement over the Non-Censored GP both in- and out-of-sample by accounting for the effect of censoring, the proposed approach is able to further improve the $R^2$ by more than 10\% on top of the Censored GP, thus underscoring the importance of accounting for the process of transfer/substitutions when building models of the aggregated true demand for products/services for which reasonable alternatives exist. The learned value of $\ell_{\text{diff}}$ converged to a value of approx.~50.6m, which, albeit a bit conservative, is still a reasonable value of how much people are willing to drive to find the nearest charging station with available chargers. Fixing $\ell_{\text{diff}} = 100$m further led to a slight improvement.

%\begin{table*}[!h]
%    \centering
%    \small
%    \begin{tabular}{l | rrrr | rrrr}
%        \toprule
%	& \multicolumn{4}{c}{\textbf{Trainset}} & \multicolumn{4}{|c}{\textbf{Testset}} \\
%         \textbf{Model} & \textbf{NLPD} & \textbf{RMSE}  & $\textbf{R}^2$ & $\textbf{RMSE (func)}$ & \textbf{NLPD} & \textbf{RMSE} & $\textbf{R}^2$ & $\textbf{RMSE (func)}$ \\ 
%        \midrule
%         True GP (Oracle) &  &  &  &  &  &  &  & \\
%         GP &  &  &  &  &  &  &  & \\
%         Censored GP &  &  &  &  &  &  &  & \\
%         Diffused Censored GP (fixed) &  &  &  &  &  &  &  & \\
%         Diffused Censored GP (learned) & \textbf{} & \textbf{} & \textbf{} & \textbf{} & \textbf{} & \textbf{} & \textbf{} & \textbf{}\\
%        \bottomrule
%    \end{tabular}
%    \caption{Template1}
%    \label{tab:template1}
%\end{table*}
%
%\begin{table*}[!h]
%    \centering
%    \small
%    \begin{tabular}{l | rrr | rrr}
%        \toprule
%	& \multicolumn{3}{c}{\textbf{Trainset}} & \multicolumn{3}{|c}{\textbf{Testset}} \\
%         \textbf{Model} & \textbf{NLPD} & \textbf{RMSE}  & $\textbf{R}^2$ & \textbf{NLPD} & \textbf{RMSE} & $\textbf{R}^2$ \\ 
%        \midrule
%         True GP (Oracle) &  &  &  &  &  & \\
%         GP &  &  &  &  &  & \\
%         Censored GP &  &  &  &  &  & \\
%         Diffused Censored GP (fixed) &  &  &  &  &  & \\
%         Diffused Censored GP (learned) & \textbf{} & \textbf{} & \textbf{} & \textbf{} & \textbf{} & \textbf{}\\
%        \bottomrule
%    \end{tabular}
%    \caption{Template2}
%    \label{tab:template2}
%\end{table*}

\section{Conclusion}
\label{sec:conclusion}

This paper proposed Diffusion-aware Censored Demand Models, which combine a Tobit likelihood with a graph diffusion process in order to model the latent process of transfer of unsatisfied demand between similar products or services. Leveraging a GP framework, we showed that our proposed Diffusion-aware Censored GP is able to better recover the (latent) true demand based on the observations of the satisfied demand and produce more accurate out-of-sample predictions. Our experimental results, based on real-world datasets for supermarket sales, bike-sharing demand, and EV charging demand, underscore the broad applicability and potential of our proposed framework. In future work, we will explore even larger-scale applications by leveraging sparse GPs.

%% The file named.bst is a bibliography style file for BibTeX 0.99c
\bibliographystyle{named}
\bibliography{ijcai25}

\begin{thebibliography}{}

\bibitem[\protect\citeauthoryear{Amemiya}{1984}]{amemiya1984tobit}
Takeshi Amemiya.
\newblock Tobit models: A survey.
\newblock {\em Journal of econometrics}, 24(1-2):3--61, 1984.

\bibitem[\protect\citeauthoryear{Basson \bgroup \em et al.\egroup
  }{2023}]{basson2023variational}
Marno Basson, Tobias~M Louw, and Theresa~R Smith.
\newblock Variational tobit gaussian process regression.
\newblock {\em Statistics and Computing}, 33(3):64, 2023.

\bibitem[\protect\citeauthoryear{Breen}{1996}]{breen1996regression}
Richard Breen.
\newblock {\em Regression models: Censored, sample selected, or truncated
  data}.
\newblock Number 111. Sage, 1996.

\bibitem[\protect\citeauthoryear{Chang \bgroup \em et al.\egroup
  }{2020}]{chang2020fast}
Paul~E Chang, William~J Wilkinson, Mohammad~Emtiyaz Khan, and Arno Solin.
\newblock Fast variational learning in state-space gaussian process models.
\newblock In {\em 2020 ieee 30th international workshop on machine learning for
  signal processing (mlsp)}, pages 1--6. IEEE, 2020.

\bibitem[\protect\citeauthoryear{Chib}{1992}]{chib1992bayes}
Siddhartha Chib.
\newblock Bayes inference in the tobit censored regression model.
\newblock {\em Journal of Econometrics}, 51(1-2):79--99, 1992.

\bibitem[\protect\citeauthoryear{DeMaris}{2004}]{demaris2004regression}
Alfred DeMaris.
\newblock {\em Regression with social data: Modeling continuous and limited
  response variables}.
\newblock John Wiley \& Sons, 2004.

\bibitem[\protect\citeauthoryear{Domarchi and
  Cherchi}{2023}]{domarchi2023electric}
Cristian Domarchi and Elisabetta Cherchi.
\newblock Electric vehicle forecasts: a review of models and methods including
  diffusion and substitution effects.
\newblock {\em Transport reviews}, 43(6):1118--1143, 2023.

\bibitem[\protect\citeauthoryear{Gammelli \bgroup \em et al.\egroup
  }{2020}]{gammelli2020estimating}
Daniele Gammelli, Inon Peled, Filipe Rodrigues, Dario Pacino, Haci~A Kurtaran,
  and Francisco~C Pereira.
\newblock Estimating latent demand of shared mobility through censored gaussian
  processes.
\newblock {\em Transportation Research Part C: Emerging Technologies},
  120:102775, 2020.

\bibitem[\protect\citeauthoryear{Gammelli \bgroup \em et al.\egroup
  }{2022}]{gammelli2022generalized}
Daniele Gammelli, Kasper~Pryds Rolsted, Dario Pacino, and Filipe Rodrigues.
\newblock Generalized multi-output gaussian process censored regression.
\newblock {\em Pattern Recognition}, 129:108751, 2022.

\bibitem[\protect\citeauthoryear{Greene}{2003}]{greene2003econometric}
William~H Greene.
\newblock Econometric analysis.
\newblock {\em Pretence Hall}, 2003.

\bibitem[\protect\citeauthoryear{Groot and Lucas}{2012}]{groot2012gaussian}
Perry Groot and Peter~JF Lucas.
\newblock Gaussian process regression with censored data using expectation
  propagation.
\newblock In {\em Proceedings of the Sixth European Workshop on Probabilistic
  Graphical Models}, pages 159--164, 2012.

\bibitem[\protect\citeauthoryear{Hamelijnck \bgroup \em et al.\egroup
  }{2021}]{hamelijnck2021spatio}
Oliver Hamelijnck, William Wilkinson, Niki Loppi, Arno Solin, and Theodoros
  Damoulas.
\newblock Spatio-temporal variational gaussian processes.
\newblock {\em Advances in Neural Information Processing Systems},
  34:23621--23633, 2021.

\bibitem[\protect\citeauthoryear{Heien and Wesseils}{1990}]{heien1990demand}
Dale Heien and Cathy~Roheim Wesseils.
\newblock Demand systems estimation with microdata: a censored regression
  approach.
\newblock {\em Journal of Business \& Economic Statistics}, 8(3):365--371,
  1990.

\bibitem[\protect\citeauthoryear{Hipolito \bgroup \em et al.\egroup
  }{2022}]{hipolito2022charging}
F~Hipolito, CA~Vandet, and J~Rich.
\newblock Charging, steady-state soc and energy storage distributions for ev
  fleets.
\newblock {\em Applied Energy}, 317:119065, 2022.

\bibitem[\protect\citeauthoryear{H{\"u}ttel \bgroup \em et al.\egroup
  }{2022}]{huttel2022modeling}
Frederik~Boe H{\"u}ttel, Inon Peled, Filipe Rodrigues, and Francisco~C Pereira.
\newblock Modeling censored mobility demand through censored quantile
  regression neural networks.
\newblock {\em IEEE Transactions on Intelligent Transportation Systems},
  23(11):21753--21765, 2022.

\bibitem[\protect\citeauthoryear{H{\"u}ttel \bgroup \em et al.\egroup
  }{2023}]{huttel2023mind}
Frederik~Boe H{\"u}ttel, Filipe Rodrigues, and Francisco~C{\^a}mara Pereira.
\newblock Mind the gap: Modelling difference between censored and uncensored
  electric vehicle charging demand.
\newblock {\em Transportation Research Part C: Emerging Technologies},
  153:104189, 2023.

\bibitem[\protect\citeauthoryear{Khan and Lin}{2017}]{khan2017conjugate}
Mohammad Khan and Wu~Lin.
\newblock Conjugate-computation variational inference: Converting variational
  inference in non-conjugate models to inferences in conjugate models.
\newblock In {\em Artificial Intelligence and Statistics}, pages 878--887.
  PMLR, 2017.

\bibitem[\protect\citeauthoryear{Li and Wang}{2003}]{li2003empirical}
Gang Li and Qi-Hua Wang.
\newblock Empirical likelihood regression analysis for right censored data.
\newblock {\em Statistica Sinica}, pages 51--68, 2003.

\bibitem[\protect\citeauthoryear{Liu \bgroup \em et al.\egroup
  }{2016}]{liu2016rebalancing}
Junming Liu, Leilei Sun, Weiwei Chen, and Hui Xiong.
\newblock Rebalancing bike sharing systems: A multi-source data smart
  optimization.
\newblock In {\em Proceedings of the 22nd ACM SIGKDD international conference
  on knowledge discovery and data mining}, pages 1005--1014, 2016.

\bibitem[\protect\citeauthoryear{Liu \bgroup \em et al.\egroup
  }{2017}]{liu2017functional}
Junming Liu, Leilei Sun, Qiao Li, Jingci Ming, Yanchi Liu, and Hui Xiong.
\newblock Functional zone based hierarchical demand prediction for bike system
  expansion.
\newblock In {\em Proceedings of the 23rd ACM SIGKDD international conference
  on knowledge discovery and data mining}, pages 957--966, 2017.

\bibitem[\protect\citeauthoryear{McDonald and Moffitt}{1980}]{mcdonald1980uses}
John~F McDonald and Robert~A Moffitt.
\newblock The uses of tobit analysis.
\newblock {\em The review of economics and statistics}, pages 318--321, 1980.

\bibitem[\protect\citeauthoryear{Schaafsma and
  Brouwer}{2020}]{schaafsma2020substitution}
Marije Schaafsma and Roy Brouwer.
\newblock Substitution effects in spatial discrete choice experiments.
\newblock {\em Environmental and Resource Economics}, 75(2):323--349, 2020.

\bibitem[\protect\citeauthoryear{Schmee and Hahn}{1979}]{schmee1979simple}
Josef Schmee and Gerald~J Hahn.
\newblock A simple method for regression analysis with censored data.
\newblock {\em Technometrics}, 21(4):417--432, 1979.

\bibitem[\protect\citeauthoryear{Sup}{}]{SupermarketKaggle}
Supermarket sales data: Sales data of vegetables in supermarket.
\newblock
  \url{https://www.kaggle.com/datasets/yapwh1208/supermarket-sales-data}.
\newblock Accessed: 2024-04-23.

\bibitem[\protect\citeauthoryear{Tebbutt \bgroup \em et al.\egroup
  }{2021}]{tebbutt2021combining}
Will Tebbutt, Arno Solin, and Richard~E Turner.
\newblock Combining pseudo-point and state space approximations for
  sum-separable gaussian processes.
\newblock In {\em Uncertainty in artificial intelligence}, pages 1607--1617.
  PMLR, 2021.

\bibitem[\protect\citeauthoryear{Tobin}{1958}]{tobin1958estimation}
James Tobin.
\newblock Estimation of relationships for limited dependent variables.
\newblock {\em Econometrica: journal of the Econometric Society}, pages 24--36,
  1958.

\bibitem[\protect\citeauthoryear{Unterluggauer \bgroup \em et al.\egroup
  }{2023}]{unterluggauer2023impact}
Tim Unterluggauer, F~Hipolito, Jeppe Rich, Mattia Marinelli, and Peter~Bach
  Andersen.
\newblock Impact of cost-based smart electric vehicle charging on urban low
  voltage power distribution networks.
\newblock {\em Sustainable Energy, Grids and Networks}, 35:101085, 2023.

\bibitem[\protect\citeauthoryear{Wan \bgroup \em et al.\egroup
  }{2018}]{wan2018demand}
Mingchao Wan, Yihui Huang, Lei Zhao, Tianhu Deng, and Jan~C Fransoo.
\newblock Demand estimation under multi-store multi-product substitution in
  high density traditional retail.
\newblock {\em European Journal of Operational Research}, 266(1):99--111, 2018.

\bibitem[\protect\citeauthoryear{Xie \bgroup \em et al.\egroup
  }{2023}]{xie2023censored}
Na~Xie, Zhiheng Li, Zhidong Liu, and Shiqi Tan.
\newblock A censored semi-bandit model for resource allocation in bike sharing
  systems.
\newblock {\em Expert Systems with Applications}, 216:119447, 2023.

\end{thebibliography}

\vfill
\pagebreak

\appendix

\section{Experimental setup details}

\subsection{Artificial (independent) data}
\label{appendix:fake_indep_data}

\begin{figure}[!ht]
    \centering
    \subfigure[Dataset A]{\includegraphics[width=0.49\textwidth,trim={2.8cm 0.6cm 2cm 1cm},clip]{fake_data.pdf}} 
    \subfigure[Dataset B]{\includegraphics[width=0.49\textwidth,trim={2.8cm 0.3cm 2cm 1cm},clip]{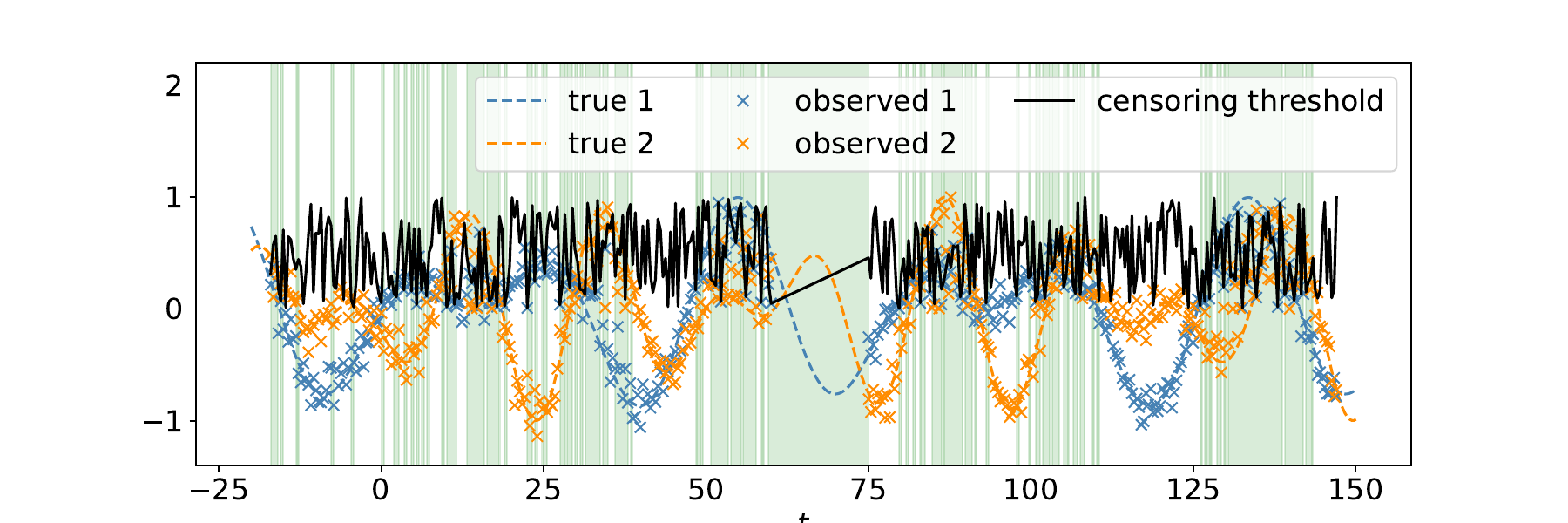}} 
    \subfigure[Dataset C]{\includegraphics[width=0.49\textwidth,trim={2.8cm 0.3cm 2cm 1cm},clip]{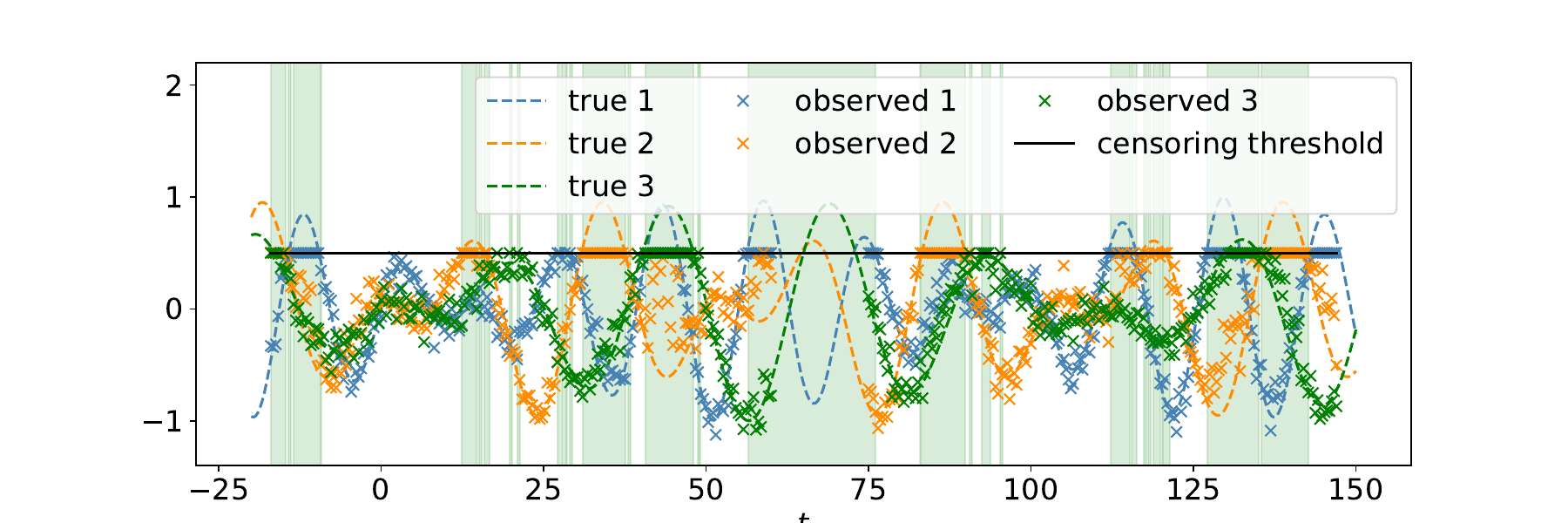}}
    \caption{Datasets A, B, and C, used for the artificial (independent) time-series experiments in Section~ \ref{subsection:artificial_data}.}
    \label{fig:indep_datasets}
\end{figure}

\noindent
\textbf{Data generation.} We generate artificial time-series by sampling from the function:
\begin{align} 
	f(t) = \cos(\theta_1 t + \theta_2 \pi + \theta_3) \times \sin(\theta_4 t + \theta_5),\nonumber
\end{align} 
with varying parameters $\bs\theta = \{\theta_1,\theta_2,\theta_3,\theta_4,\theta_5\}$, and then adding Gaussian white noise. Figure~\ref{fig:indep_datasets} shows the generated data for Datasets A, B, and C. 

\vspace{0.1cm}
\noindent
\textbf{Data sizes and splits.} We sample 400 evenly-spaced time points for training and use another 1000 time points for testing. 

\vspace{0.1cm}
\noindent
\textbf{Censoring simulation.} In Datasets A and C, we simulate the censoring process by setting a constant censoring threshold (supply). In Dataset B, the threshold is sampled randomly from a uniform distribution for each time step, thus resulting in a stochastic supply that causes very sporadic supply shortages rather than shortages lasting for longer time intervals as in Datasets A and C - see Figure~\ref{fig:indep_datasets}. 

\vspace{0.1cm}
\noindent
\textbf{Excess demand transfer simulation.} For the datasets with only 2 time-series (Datasets A and B), simulating the demand for just 2 products, the demand above the threshold is sent to the other time-series. In Dataset C we simulated time-series for 3 products. Whenever Product 1 or Product 3 are out of supply, their excess demand goes to Product 2, while the excess demand from the latter is distributed equally among Products 1 and 3. If the resulting demand for the other time-series now becomes higher than the threshold, then we throw away the excess demand. By simulating only one step of demand transfer, we emulate a scenario where customers have perfect information about the supply of all products (hence they only select an alternative once). 

\vspace{0.1cm}
\noindent
\textbf{Kernels.} All models in this experiment use the same underlying GP framework with independent GPs for each product (i.e.~$\kappa_\bp(\bp,\bp') = \mathbbm{1}(\bp = \bp')$) and $\kappa_t(t,t')$ being a Matern kernel. We initialize the likelihood variance $\sigma^2 = 0.1$. Both the hyper-parameters of $\kappa_t(t,t')$ and the likelihood variance $\sigma^2$ are optimized during training via type-II maximum likelihood estimation.

\vspace{0.1cm}
\noindent
\textbf{Diffusion inputs and hyper-parameters.} The input features $\bp$ describing the products are 2-D vectors. Namely, we use the following vectors: $\bp_1 = (-1,-1)$, $\bp_2 = (0,0)$, and $\bp_3 = (1,1)$. We initialize $\ell_{\text{diff}} = 1$ and set $N_{\text{diff}} = 1$. 

\vspace{0.1cm}
\noindent
\textbf{Training and learning rates.} We train all models for 300 iterations, observing convergence for all of them. We set the learning rate of the CVI updates to $\beta = 0.1$, and the learning rate of the Adam optimizer (used for hyper-parameter tuning) to 0.1.

\subsection{Artificial spatio-temporal data}
\label{appendix:fake_ST_data}

\begin{figure}[!ht]
    \centering
    \subfigure{\includegraphics[width=0.49\textwidth,trim={2.8cm 0cm 1cm 0cm},clip]{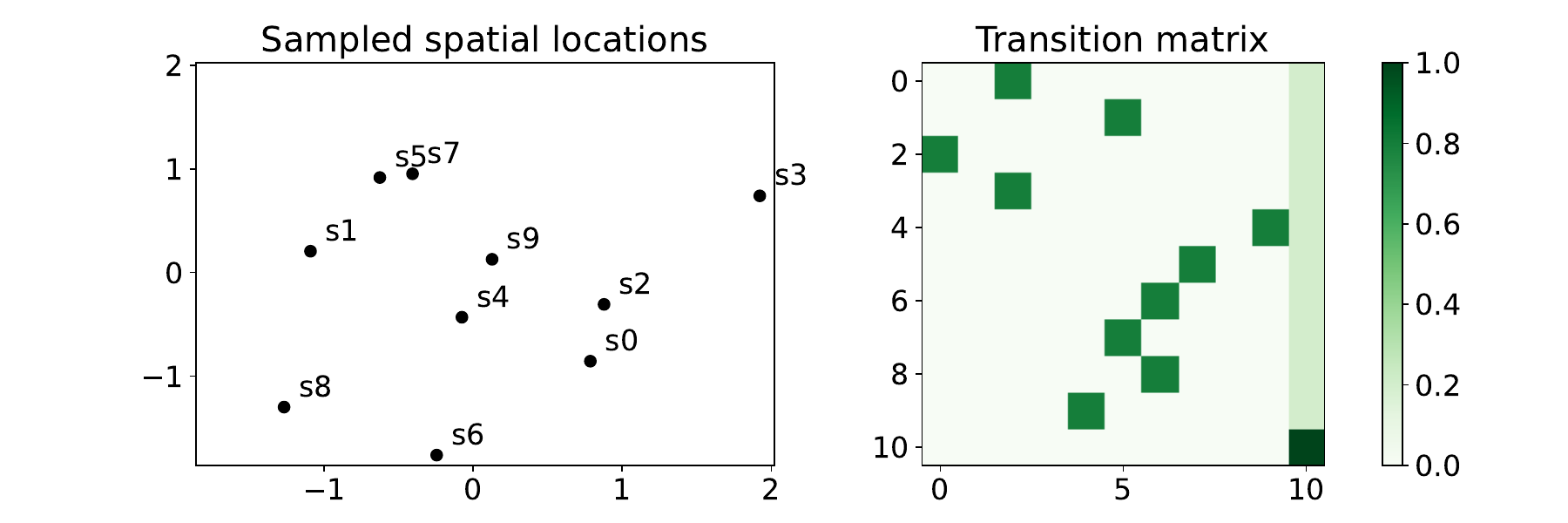}} 
    \subfigure{\includegraphics[width=0.49\textwidth,trim={2.8cm 0.0cm 2cm 0cm},clip]{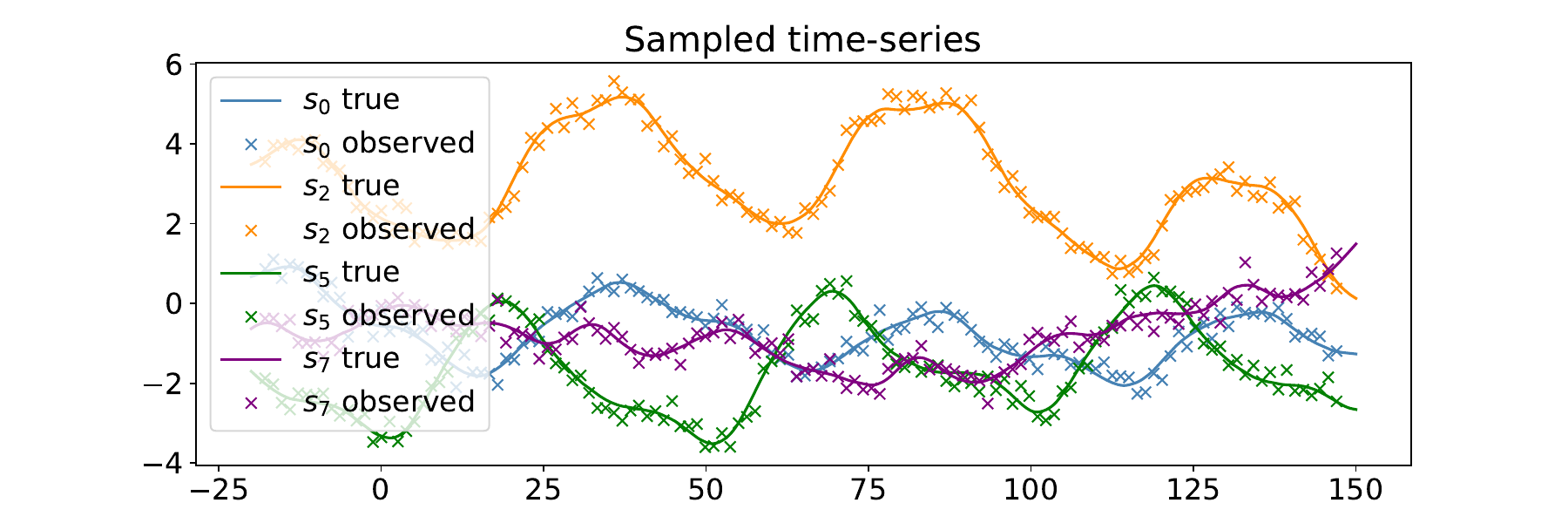}} 
    \caption{Sampled spatial location (top left), transition matrix used for the diffusion process (top right), and 4 selected time-series sampled (bottom).}
    \label{fig:st_dataset}
\end{figure}

\noindent
\textbf{Data generation.} We generate artificial time-series by sampling from a spatio-temporal GP with a separable kernel with a Periodic+Matern component over time and a Matern component over space, to which we then add Gaussian noise. The spatial 2-D locations are sampled uniformly in the range [-2,2]. We use 10 spatial locations (and, hence, we sample 10 time-series). Figure~\ref{fig:st_dataset} shows the generated data (corresponding to the true demand) for Datasets ST-1 and ST-2. 

\vspace{0.1cm}
\noindent
\textbf{Data sizes and splits.} The sampled data consists of 10 time-series ($N_p = 10$) - one per location. We use 130 evenly-spaced time points for training and another 400 time points for testing. 

\vspace{0.1cm}
\noindent
\textbf{Censoring simulation.} In this experiment, we simulate the censoring process with a 2-state Markov model, where the states represent ``with supply'' and ``out of supply'' states. The probability of transitioning from a ``with supply'' state to an ``out of supply'' state is 3\%, with the reverse transition probability being 10\%. When in an ``out of supply'' state, the observed value is capped at a constant value. 

\vspace{0.1cm}
\noindent
\textbf{Excess demand transfer simulation.} The excess demand is transferred to the nearby locations by simulating the diffusion process described in Section~\ref{subsec:transition_dynamics} with diffusion parameter $\ell_{\text{diff}} = 0.1$ and with a sink node probability of 20\%. The resulting transition matrix $\bT$ is shown in Figure~\ref{fig:st_dataset} (top right). In this scenario, the product features $\bp$ correspond to the 2-D spatial locations. We use $N_{\text{diff}}=1$ and $N_{\text{diff}}=2$ diffusion steps for datasets ``Dataset ST-1'' and ``Dataset ST-2'', respectively. 

\vspace{0.1cm}
\noindent
\textbf{Kernels.} All models use the same underlying GP framework with a spatio-temporal GP with a separable kernel with a Periodic+Matern component over time and a Matern component over space. We initialize the likelihood variance $\sigma^2 = 0.1$. Both the hyper-parameters of $\kappa_t(t,t')$ and $\kappa_{\bp}(\bp,\bp')$, as well as the likelihood variance $\sigma^2$ are optimized during training via type-II maximum likelihood estimation. 

\vspace{0.1cm}
\noindent
\textbf{Diffusion inputs and hyper-parameters.} In this experiment, the input features $\bp$ describing the products correspond to the 2-D spatial locations. We initialize $\ell_{\text{diff}} = 0.1$ and set $\pi_{\text{diff}} = 0.2$. 

\vspace{0.1cm}
\noindent
\textbf{Training and learning rates.} The models were trained for 5000 iterations, observing convergence for all of them. We set the learning rate of the CVI updates to $\beta = 1$, and the learning rate of the Adam optimizer (used for hyper-parameter tuning) to 0.1.

\subsection{Supermarket sales}
\label{appendix:sales_details}

\noindent
\textbf{Data sizes and splits.} The dataset consists of 3 years' worth of supermarket sales data between 2020 and 2023, which we aggregate on a daily basis. Using a 90/10\% train/test split resulted in 964 observations for training and 107 for testing. 

\vspace{0.1cm}
\noindent
\textbf{Censoring simulation.} In this experiment, we simulate the censoring process by using a 2-state Markov model, where the states represent ``with supply'' and ``out of supply'' states. The probability of transitioning from a ``with supply'' state to an ``out of supply'' state is 2\%, with the reverse transition probability being 5\%. We assume that the dataset observations correspond to the true demand $f_t$. When an ``out of supply'' state is entered at time $t$, the available supply $u_t$ is sampled uniformly as $u_t \sim \text{Uniform}(0, f_t / 2)$, and the observed (censored) value is set to that value: $y_t = u_t$. For as long as it remains in an ``out of supply'' state, the observed value is set to that value. Upon recovery to a ``with supply'' state, the observed value becomes again the true demand ($y_t = f_t$). The intuition behind this process is to mimic continuous periods where supply is limited. Applying this censoring simulation process resulted in 11.9\% and 18.5\% censored observations for time-series 1 and 2, respectively. 

\vspace{0.1cm}
\noindent
\textbf{Excess demand transfer simulation.} The excess demand is transferred in its entirety to the other time-series, thus simulating a scenario where customers always go for an alternative (if available) when their preferred product is out of supply. If the resulting demand for the other time-series now becomes higher than the threshold, then we throw away the excess demand. By simulating only one step of demand transfer, we emulate a scenario where customers have perfect information about the supply of all products (hence they only select an alternative once). 

\vspace{0.1cm}
\noindent
\textbf{Kernels.} All models use the same underlying GP framework with independent GPs for each product (i.e.~$\kappa_\bp(\bp,\bp') = \mathbbm{1}(\bp = \bp')$) and $\kappa_t(t,t')$ being a Matern kernel. We initialize the likelihood variance $\sigma^2 = 1$. Both the hyper-parameters of $\kappa_t(t,t')$ and the likelihood variance $\sigma^2$ are optimized during training via type-II maximum likelihood estimation.

\vspace{0.1cm}
\noindent
\textbf{Diffusion inputs and hyper-parameters.} The input features $\bp$ describing the products are 2-D vectors. We initialize $\ell_{\text{diff}} = 1$ and set $N_{\text{diff}} = 1$. 

\vspace{0.1cm}
\noindent
\textbf{Training and learning rates.} We train all models for 2000 iterations, observing convergence for all of them. We set the learning rate of the CVI updates to $\beta = 0.1$, and the learning rate of the Adam optimizer (used for hyper-parameter tuning) to 0.01.

\subsection{Bike sharing}
\label{appendix:bike_details}

\noindent
\textbf{Data sizes and splits.} The data used for this experiment consists of 2 months (June and July of 2018) of hourly bike-sharing demand for 4 bicycle hubs. Using a 70/30\% train/test split resulted in 1026 observations for training and 439 for testing. 

\begin{table*}[!th]
    \centering
    \small
    \begin{tabular}{cl | rrrr | rrrr}
        \toprule
	&& \multicolumn{4}{c}{\textbf{Trainset}} & \multicolumn{4}{|c}{\textbf{Testset}} \\
	&\textbf{Model} & \textbf{NLPD} & \textbf{RMSE}  & $\textbf{R}^2$ & $\textbf{RMSE funct.}$ & \textbf{NLPD} & \textbf{RMSE} & $\textbf{R}^2$ & $\textbf{RMSE funct.}$ \\ 
        \midrule
         \multirow{5}{*}{\rotatebox[origin=c]{90}{Dataset A}} & True GP (Oracle) & -1.850 & 0.095 & 0.964 & 0.029 & -1.771 & 0.099 & 0.960 & 0.029\\
         & (Non-Censored) GP & 0.634 & 0.183 & 0.867 & 0.163 & 0.853 & 0.188 & 0.858 & 0.163\\
         & Censored GP & -0.795 & 0.155 & 0.904 & 0.130 & -0.674 & 0.162 & 0.895 & 0.130\\
         %& Diffused Censored GP (fixed) & \textbf{-0.800} & {0.103} & {0.957} & {0.051} & \textbf{-0.671} & {0.108} & {0.953} & {0.051}\\
         & Diffused Censored GP (fixed) & {-0.816} & \textbf{0.103} & \textbf{0.958} & {0.051} & {-0.685} & \textbf{0.108} & \textbf{0.953} & \textbf{0.051}\\
         %& Diffused Censored GP (learned) & {-0.759} & \textbf{0.114} & \textbf{0.948} & \textbf{0.073} & {-0.678} & \textbf{0.119} & \textbf{0.943} & \textbf{0.073}\\
         & Diffused Censored GP (learned) & \textbf{-0.859} & {0.112} & {0.950} & {0.070} & \textbf{-0.751} & {0.118} & {0.944} & {0.071}\\
        \midrule
         \multirow{5}{*}{\rotatebox[origin=c]{90}{Dataset B}} & True GP (Oracle) & -1.850 & 0.095 & 0.964 & 0.029 & -1.701 & 0.103 & 0.957 & 0.029\\
         & (Non-Censored) GP & -0.594 & 0.180 & 0.870 & 0.158 & -0.505 & 0.188 & 0.857 & 0.158\\
         & Censored GP & \textbf{-1.063} & 0.144 & 0.918 & 0.111 & \textbf{-1.004} & 0.148 & 0.912 & 0.111\\
         %& Diffused Censored GP (fixed) & -0.109 & \textbf{0.101} & \textbf{0.959} & \textbf{0.039} & 0.956 & \textbf{0.107} & \textbf{0.953} & \textbf{0.039}\\
         & Diffused Censored GP (fixed) & 0.103 & {0.101} & \textbf{0.960} & {0.038} & 0.911 & {0.107} & \textbf{0.954} & {0.038}\\
         %& Diffused Censored GP (learned) & {-0.119} & {0.101} & {0.960} & {0.037} & {0.630} & {0.107} & {0.954} & {0.037}\\
         & Diffused Censored GP (learned) & {-0.131} & \textbf{0.100} & \textbf{0.960} & \textbf{0.037} & {0.590} & \textbf{0.106} & \textbf{0.954} & \textbf{0.037}\\
        \midrule
         \multirow{5}{*}{\rotatebox[origin=c]{90}{Dataset C}} & True GP (Oracle) & -2.871 & 0.092 & 0.964 & 0.032 & -2.527 & 0.104 & 0.954 & 0.031\\
         & (Non-Censored) GP & 0.545 & 0.184 & 0.864 & 0.162 & 0.805 & 0.188 & 0.856 & 0.162\\
         & Censored GP & -1.527 & 0.148 & 0.912 & 0.123 & -1.224 & 0.157 & 0.900 & 0.122\\
         %& Diffused Censored GP (fixed) & -1.540 & \textbf{0.099} & \textbf{0.959} & \textbf{0.043} & -1.312 & \textbf{0.107} & \textbf{0.952} & \textbf{0.043}\\
         & Diffused Censored GP (fixed) & -1.533 & \textbf{0.099} & \textbf{0.959} & \textbf{0.043} & -1.311 & \textbf{0.107} & \textbf{0.952} & \textbf{0.042}\\
         %& Diffused Censored GP (learned) & \textbf{-1.500} & \textbf{0.117} & \textbf{0.944} & {0.078} & \textbf{-1.228} & {0.125} & {0.935} & {0.078}\\
         & Diffused Censored GP (learned) & \textbf{-1.609} & {0.122} & {0.940} & {0.086} & \textbf{-1.380} & {0.130} & {0.930} & {0.086}\\
        \bottomrule
    \end{tabular}
    \caption{Complete version of the experimental results (incl.~train set performance) obtained for the three artificial datasets of independently-generated (uncorrelated) time-series. ``RMSE funct.'' further measures the RMSE to the true underlying (noiseless) function.} 
    %Inclides: paper - demo\_fakeIndepData1\_indepKernel\_throwExcess.ipynb, paper - demo\_fakeIndepData2\_indepKernel\_throwExcess.ipynb, paper - demo\_fakeIndepData3\_indepKernel\_throwExcess.ipynb}
    \label{tab:fakeIndepData_indepKernel_complete}
\end{table*}

\vspace{0.1cm}
\noindent
\textbf{Censoring simulation.} We employ the same procedure as proposed by \citet{gammelli2020estimating}. Since the original dataset contains both bicycle pick-ups and drop-offs, one can use the difference between the two as a proxy for the available supply. Therefore, if one assumes an initial supply at the beginning of each day (it is common practice for bike-sharing operators to manually reposition the bicycles in their fleets during the night in preparation for the next day), then one can keep track of the available supply throughout the rest of the day using information about the pick-ups and drop-offs. We use this estimate of the supply to then censor the time-series of pickups. The amount of censoring is then fully dependent on the assumed initial supply value at the beginning of each day, which we randomly sample from a uniform distribution. Applying this censoring simulation process resulted in 20.5\%, 3.4\%, 9.1\%, and 2.0\% of censored observations for time-series 1 to 4, respectively. 

\vspace{0.1cm}
\noindent
\textbf{Excess demand transfer simulation.} The excess demand is transferred to the nearby locations by simulating the diffusion process described in Section~\ref{subsec:transition_dynamics} with diffusion parameter $\ell_{\text{diff}} = 0.03$. In this scenario, the product features $\bp$ correspond to the 2-D spatial locations. We apply one single diffusion step ($N_{\text{diff}}=1$). If the resulting demand for the other time-series now becomes higher than the threshold, then we throw away the excess demand. Therefore, this scenario entails the assumption that when a bicycle hub is out of supply, the users will always try the nearest hub (usually within walking distance for most cities), and if that is out of supply as well, then they give up (e.g., they would use public transport or an alternative transportation mode instead). 

\vspace{0.1cm}
\noindent
\textbf{Kernels.} All models use the same underlying GP framework with a spatio-temporal GP with a separable kernel with a Periodic+Matern component over time and a Matern component over space. We initialize the likelihood variance $\sigma^2 = 1$. Both the hyper-parameters of $\kappa_{\bp}(\bp,\bp')$ and the likelihood variance $\sigma^2$ are optimized during training via type-II maximum likelihood estimation. 

\vspace{0.1cm}
\noindent
\textbf{Diffusion inputs and hyper-parameters.} In this experiment, the input features $\bp$ describing the products correspond to the 2-D spatial locations of the hubs. For the fixed version of the proposed approach (``DCGP-f''), we set $\ell_{\text{diff}} = 0.03$ (true value). 

\vspace{0.1cm}
\noindent
\textbf{Training and learning rates.} We train all models for 1000 iterations, observing convergence for all of them. We set the learning rate of the CVI updates to $\beta = 0.1$, and the learning rate of the Adam optimizer (used for hyper-parameter tuning) to 0.1.

\subsection{EV charging}
\label{appendix:EV_charging_details} 

\begin{figure}
    \centering
    \includegraphics[width=.5\textwidth,trim={4.0cm 2.6cm 1.8cm 3.0cm},clip]{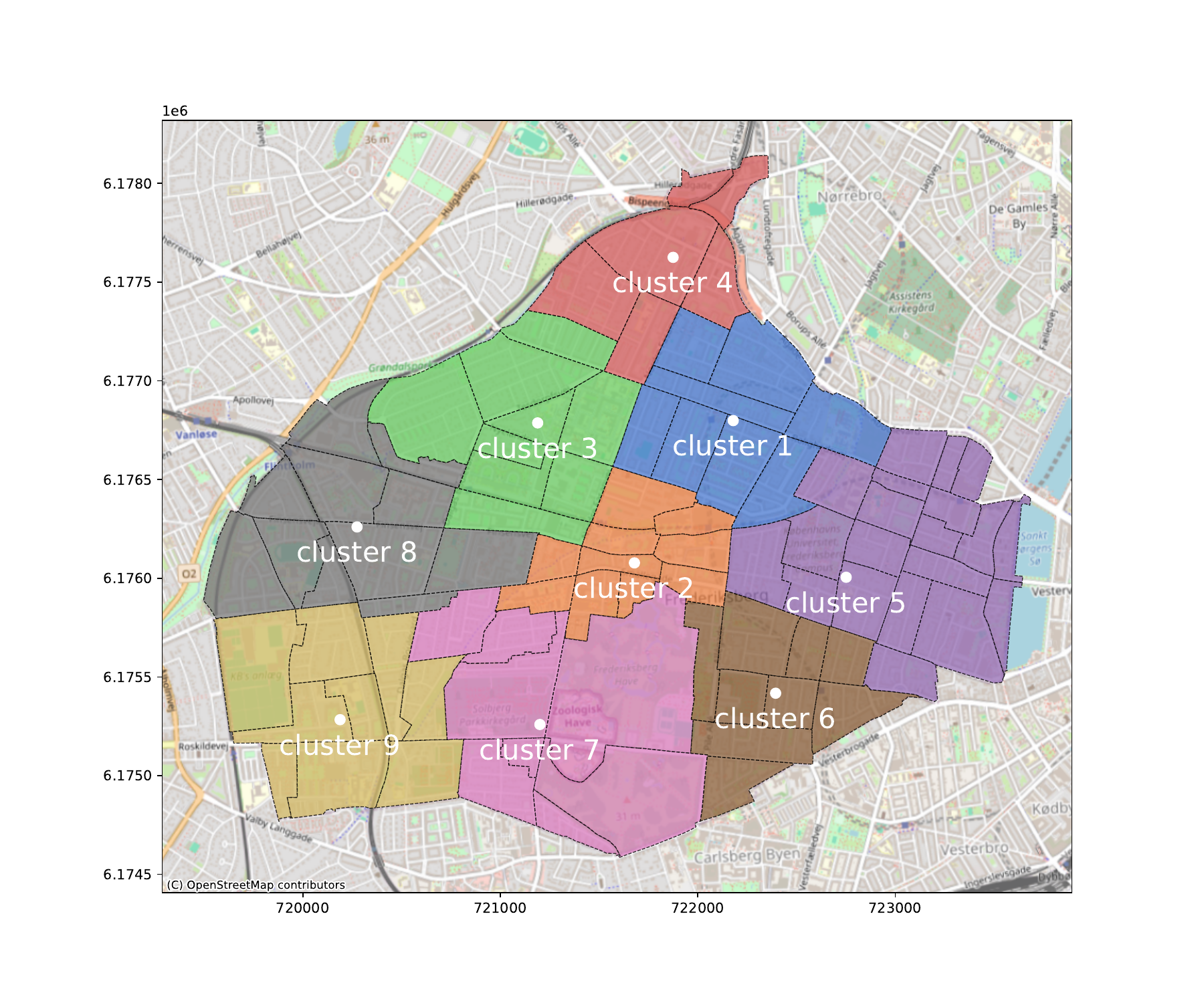}
    \caption{Map of the Frederiksberg commune in Copenhagen (green shaded area), which we partition into 9 sub-areas (clusters) whose centroids are represented by the purple points.}
    \label{fig:clustering_frederiksberg}
\end{figure}

\noindent
\textbf{Data sizes and splits.} The data used for this experiment consists of 4 weeks of data of EV charging events generated by the agent-based simulation tool, GAIA \citep{unterluggauer2023impact} for the commune of Frederiksberg in Copenhagen, Denmark. We cluster Frederiksberg into 9 clusters using k-means - see Figure~\ref{fig:clustering_frederiksberg}. We aggregate the data in 5-minute intervals, thus resulting in 9 time-series of length 7776. We use 70\% of the data for training the different models and the remaining 30\% for testing. 

\vspace{0.1cm}
\noindent
\textbf{Kernels.} All models use the same underlying GP framework with a spatio-temporal GP with a separable kernel with a Periodic+Matern component over time and a Matern component over space. We initialize the likelihood variance $\sigma^2 = 1$. Both the hyper-parameters of $\kappa_{\bp}(\bp,\bp')$ and the likelihood variance $\sigma^2$ are optimized during training via type-II maximum likelihood estimation. 

\vspace{0.1cm}
\noindent
\textbf{Diffusion inputs and hyper-parameters.} In this experiment, the input features $\bp$ describing the products correspond to the 2-D spatial locations of the cluster centroids. We initialize $\ell_{\text{diff}} = 10$. 

\vspace{0.1cm}
\noindent
\textbf{Training and learning rates.} We train all models for 1000 iterations, observing convergence for all of them. We set the learning rate of the CVI updates to $\beta = 1$, and the learning rate of the Adam optimizer (used for hyper-parameter tuning) to 0.1.

\section{Addtional experiments}

\subsection{Analysis of the inferred true demand}

\begin{figure*}[!th]
    \centering
    \subfigure[(Non-Censored) GP posterior]{\includegraphics[width=0.49\textwidth,trim={2.8cm 0.3cm 2cm 1.2cm},clip]{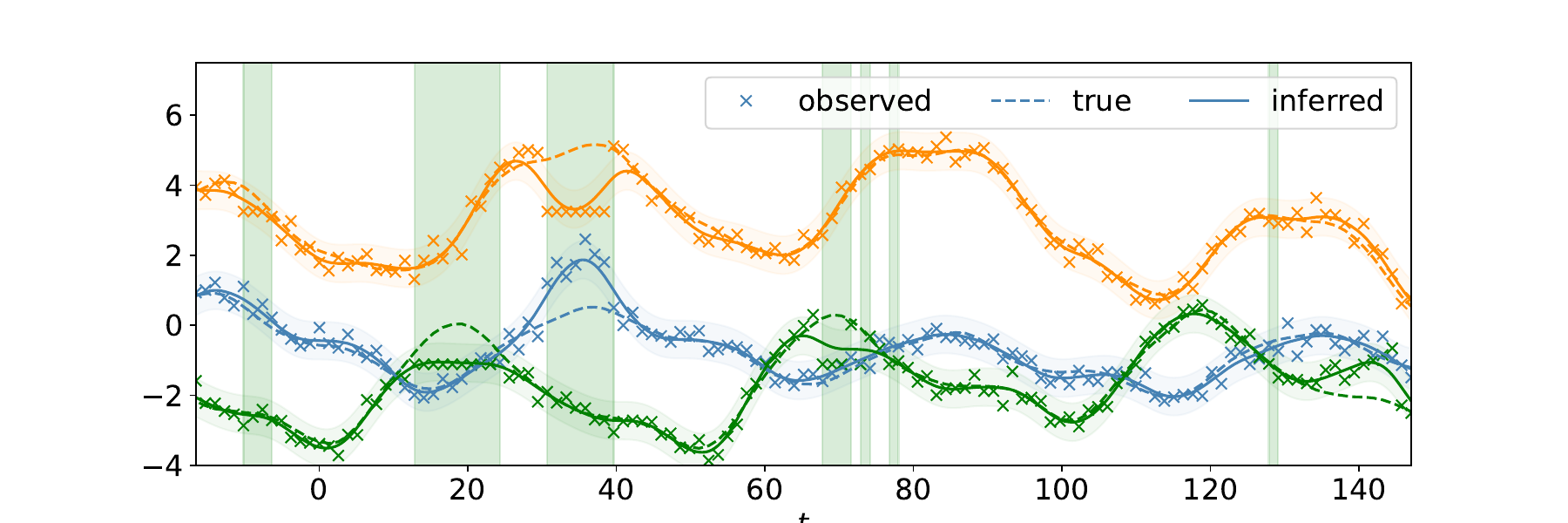}} 
    \subfigure[Censored GP posterior]{\includegraphics[width=0.49\textwidth,trim={2.8cm 0.3cm 2cm 1.2cm},clip]{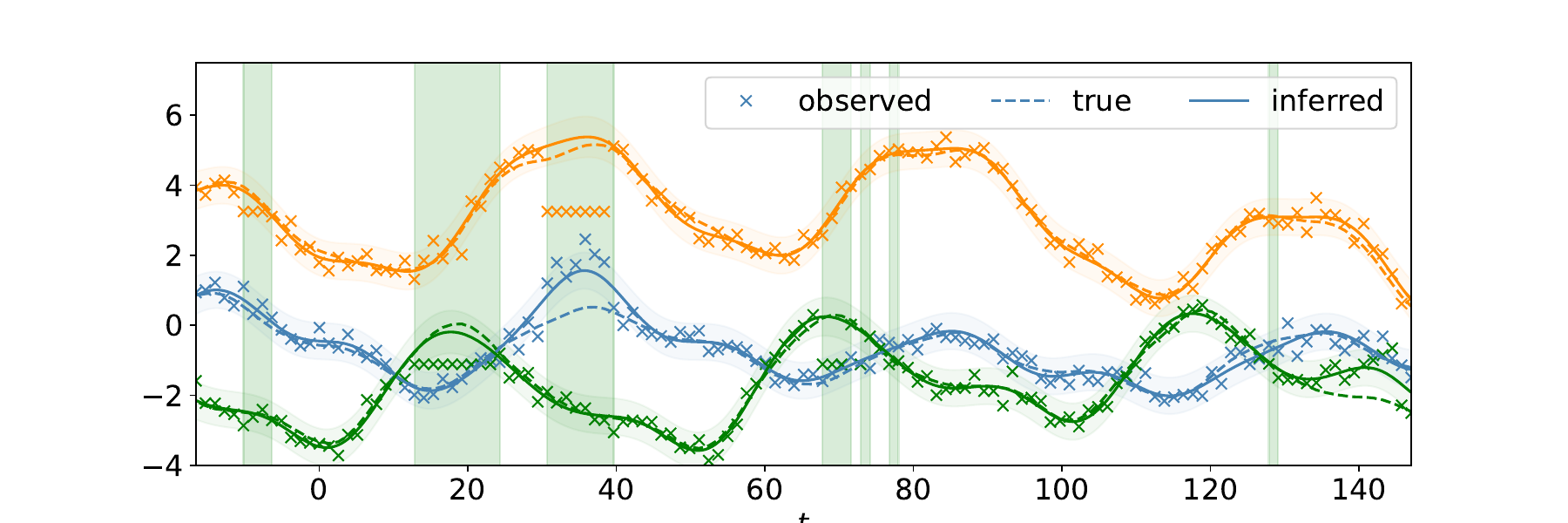}} \\ \vspace{-0.2cm}
    \subfigure[Diffused Censored GP (fixed) posterior]{\includegraphics[width=0.49\textwidth,trim={2.8cm 0.3cm 2cm 1.2cm},clip]{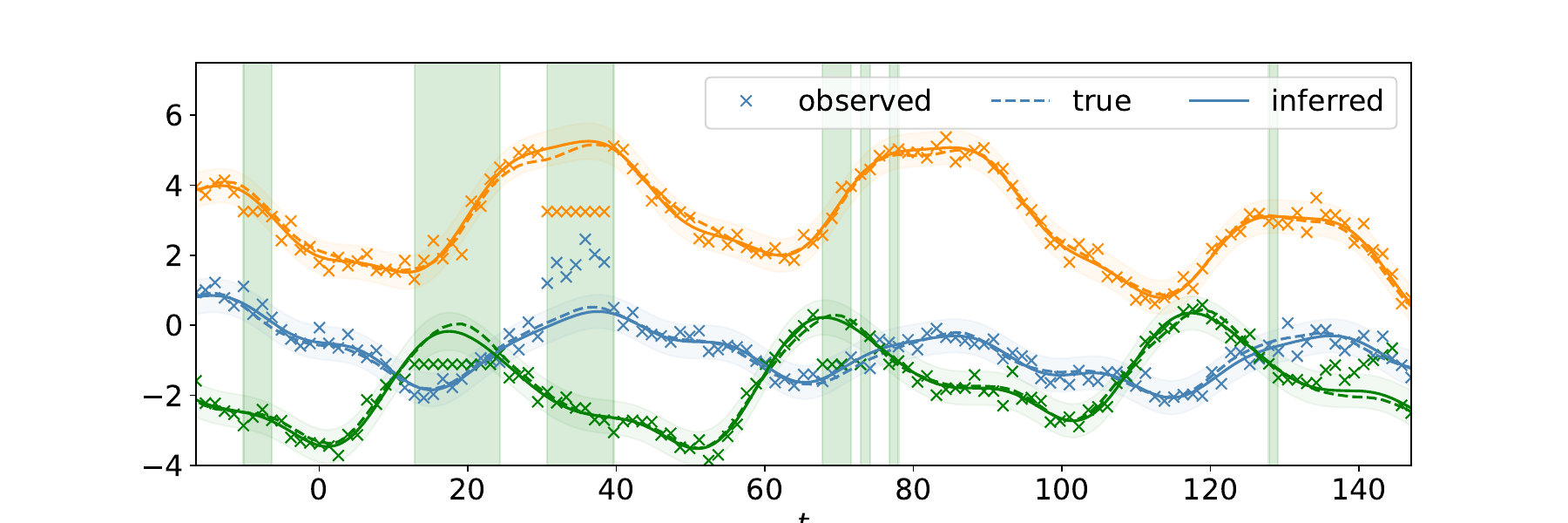}}
    \subfigure[Diffused Censored GP (learned) posterior]{\includegraphics[width=0.49\textwidth,trim={2.8cm 0.3cm 2cm 1.2cm},clip]{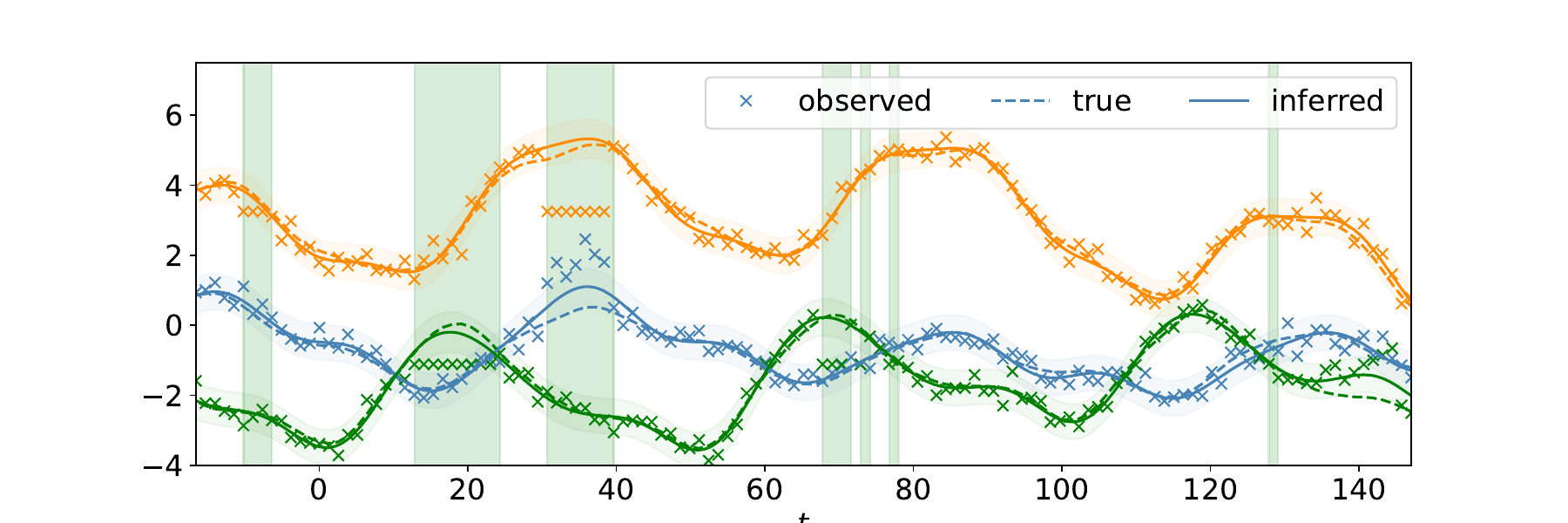}}\vspace{-0.3cm}
    \caption{GP posteriors over the true demand $\bff_n$ inferred by the different approaches for 3 of the 10 artificially-sampled time-series in Dataset ST-1. The green-shaded areas indicate time indexes where at least one of the time-series is subject to censoring. As shown in (c) and (d), the proposed approach achieves the best approximation to the true demand.}
    \label{fig:fits_to_STdata}
\end{figure*}
\label{appendix:analysis_inferred_demand}

\begin{table*}[!th]
    \centering
    \small
    \begin{tabular}{cl | rrrr | rrrr}
        \toprule
	&& \multicolumn{4}{c}{\textbf{Trainset}} & \multicolumn{4}{|c}{\textbf{Testset}} \\
	&\textbf{Model} & \textbf{NLPD} & \textbf{RMSE}  & $\textbf{R}^2$ & $\textbf{RMSE funct.}$ & \textbf{NLPD} & \textbf{RMSE} & $\textbf{R}^2$ & $\textbf{RMSE funct.}$ \\ 
        \midrule
         \multirow{5}{*}{\rotatebox[origin=c]{90}{Dataset ST-1}} & True GP (Oracle) & -0.094 & 0.239 & 0.937 & 0.144 & -0.031 & 0.273 & 0.924 & 0.178\\
         & (Non-Censored) GP & 0.220 & 0.325 & 0.893 & 0.286 & 0.327 & 0.368 & 0.869 & 0.313\\
         & Censored GP & 0.043 & 0.276 & 0.915 & 0.207 & 0.111 & 0.313 & 0.900 & 0.239\\
         & Diffused Censored GP (fixed) & \textbf{-0.084} & \textbf{0.238} & \textbf{0.938} & \textbf{0.143} & \textbf{-0.019} & \textbf{0.268} & \textbf{0.926} & \textbf{0.172}\\
         & Diffused Censored GP (learned) & {-0.015} & {0.251} & {0.931} & {0.167} & {0.047} & {0.281} & {0.920} & {0.191}\\
        \midrule
         \multirow{7}{*}{\rotatebox[origin=c]{90}{Dataset ST-2}} & True GP (Oracle) & -0.094 & 0.239 & 0.937 & 0.144 & -0.031 & 0.273 & 0.924 & 0.178\\
         & (Non-Censored) GP & 0.188 & 0.315 & 0.900 & 0.272 & 0.285 & 0.358 & 0.877 & 0.300\\
         & Censored GP & 0.064 & 0.280 & 0.916 & 0.216 & 0.140 & 0.318 & 0.899 & 0.246\\
         & Diffused Censored GP (1-step, fixed) & \textbf{-0.044} & 0.249 & 0.934 & 0.163 & {0.020} & 0.282 & 0.920 & 0.193\\
         & Diffused Censored GP (1-step, learnable) & {0.016} & {0.266} & {0.925} & {0.193} & {0.088} & {0.302} & {0.909} & {0.223}\\
         & Diffused Censored GP (2-step, fixed) & {-0.044} & \textbf{0.242} & \textbf{0.937} & \textbf{0.150} & \textbf{0.018} & \textbf{0.269} & \textbf{0.926} & \textbf{0.172}\\
         & Diffused Censored GP (2-step, learnable) & -0.005 & 0.255 & 0.931 & 0.174 & 0.060 & 0.286 & 0.918 & 0.199\\
        \bottomrule
    \end{tabular}
    \caption{Complete version of the experimental results (incl.~train set performance) obtained for the two artificially-generated spatio-temporal datasets. ``RMSE funct.'' further measures the RMSE to the true underlying (noiseless) function.}
    %Includes: paper - demo\_fakeSTdata9\_STkernel.ipynb, paper - demo\_fakeSTdata10\_STkernel.ipynb}
    \label{tab:fakeSTdata_STkernel_complete}
\end{table*}

\textbf{Indepedent time-series.} Table~\ref{tab:fakeIndepData_indepKernel_complete} shows the complete set of results for the artificially generated datasets (Datasets A, B, and C) described in Section~\ref{subsection:artificial_data}. The results show consistency between the train set and test set performance, with a lower error in recovering the true demand for the train set being translated into better performance in the test set. Interestingly, for Datasets A and C, the best results were obtained with a fixed $\ell_{\text{diff}} = 1$, while for Dataset B learning $\ell_{\text{diff}}$ resulted in slightly better results, which we attribute to the fact that the sporadic supply shortages present in this dataset constitute a simpler problem, while in Datasets A and C the supply shortages occur in longer bursts, thus causing the underlying GP model to be less confident in ``explaining'' them through the diffusion process. Overall, the results show that the proposed approach significantly outperforms all the other baselines, and in some cases, it is able to achieve near ``oracle'' performance (particularly, the variant with $\ell_{\text{diff}}$ fixed). 

\textbf{Spatio-temporal data.} Table~\ref{tab:fakeSTdata_STkernel_complete} shows the complete set of results for the artificially generated spatio-temporal datasets (Datasets ST-1 and ST-2) described in Section~\ref{subsection:artificial_data}. The results show consistency between the train set and test set performance and further highlight the capability of the proposed approach to infer the true demand from the observed demand and supply information. Figure~\ref{fig:fits_to_STdata} further shows the posterior distribution over the true demand $\bff_n$ inferred by the different approaches for 3 of the 10 artificially-sampled time-series in Dataset ST-1. As shown in Figure~\ref{fig:fits_to_STdata}a, the standard (Non-Censored) GP simply fits the observed values of demand, thus resulting in a biased GP posterior that does not reflect the true demand. The Censored GP (Figure~\ref{fig:fits_to_STdata}b) is able to take into account information about censoring to be ``less sensitive'' to censored observations, thus being able to infer that the true demand should be higher than observed at censored locations - see, e.g., the orange curve around time index ~30 to ~40 in the x-axis. However, it remains oblivious that the abnormally high demand observed for the blue curve around the same time interval can be explained by the censoring occurring for the orange time-series. On the other, both variants of the proposed approach (Figures~\ref{fig:fits_to_STdata}c and \ref{fig:fits_to_STdata}d) are able to account for the process of transfer of ``excess demand'', thus resulting in a more accurate approximation to the true demand.

\subsection{Sensitivity analysis to model misspecification}
\label{appendix:model_misspec}

\begin{table*}[!th]
    \centering
    \small
    \begin{tabular}{l | rrrr | rrrr}
        \toprule
	& \multicolumn{4}{c}{\textbf{Trainset}} & \multicolumn{4}{|c}{\textbf{Testset}} \\
	\textbf{Model} & \textbf{NLPD} & \textbf{RMSE}  & $\textbf{R}^2$ & $\textbf{RMSE funct.}$ & \textbf{NLPD} & \textbf{RMSE} & $\textbf{R}^2$ & $\textbf{RMSE funct.}$ \\ 
        \midrule
         Censored GP & 0.043 & 0.276 & 0.915 & 0.207 & 0.111 & 0.313 & 0.900 & 0.239\\
         \midrule
         Diffused Censored GP (fixed $\ell_{\text{diff}} = 0.05$) & \textbf{-0.084} & \textbf{0.238} & \textbf{0.938} & \textbf{0.143} & \textbf{-0.019} & \textbf{0.268} & \textbf{0.926} & \textbf{0.172}\\
         Diffused Censored GP (fixed $\ell_{\text{diff}} = 0.1$ $\leftarrow$ true) & \textbf{-0.084} & \textbf{0.238} & \textbf{0.938} & \textbf{0.143} & \textbf{-0.019} & \textbf{0.268} & \textbf{0.926} & \textbf{0.172}\\
         Diffused Censored GP (fixed $\ell_{\text{diff}} = 0.2$) & \textbf{-0.084} & \textbf{0.238} & \textbf{0.938} & \textbf{0.143} & \textbf{-0.019} & \textbf{0.269} & \textbf{0.926} & \textbf{0.172}\\
         Diffused Censored GP (fixed $\ell_{\text{diff}} = 0.5$) & {-0.050} & {0.250} & {0.933} & {0.165} & {0.016} & {0.285} & {0.919} & {0.198}\\
         Diffused Censored GP (fixed $\ell_{\text{diff}} = 1.0$) & {-0.003} & {0.263} & {0.925} & {0.186} & {0.064} & {0.299} & {0.910} & {0.219}\\
         \midrule
         Diffused Censored GP (learned $\ell_{\text{diff}}$) & {-0.015} & {0.251} & {0.931} & {0.167} & {0.047} & {0.281} & {0.920} & {0.191}\\
        \bottomrule
    \end{tabular}
    \caption{Sensitivity analysis results for model misspecification with respect to the diffusion hyper-parameter $\ell_{\text{diff}}$ (diffusion lengthscale). ``RMSE funct.'' further measures the RMSE to the true underlying (noiseless) function.}
    %Includes: paper - demo\_fakeSTdata9\_STkernel.ipynb}
    \label{tab:fakeSTdata_wrong_ell}
\end{table*}

\begin{table*}[!th]
    \centering
    \small
    \begin{tabular}{l | rrrr | rrrr}
        \toprule
	& \multicolumn{4}{c}{\textbf{Trainset}} & \multicolumn{4}{|c}{\textbf{Testset}} \\
	\textbf{Model} & \textbf{NLPD} & \textbf{RMSE}  & $\textbf{R}^2$ & $\textbf{RMSE funct.}$ & \textbf{NLPD} & \textbf{RMSE} & $\textbf{R}^2$ & $\textbf{RMSE funct.}$ \\ 
        \midrule
         Censored GP & 0.016 & 0.284 & 0.913 & 0.239 & 0.141 & 0.341 & 0.884 & 0.281\\
         \midrule
         Diffused Censored GP (fixed $\pi_{\text{diff}} = 0.0$) & {-0.066} & {0.239} & {0.936} & {0.142} & {-0.002} & {0.267} & {0.926} & {0.169}\\
         Diffused Censored GP (fixed $\pi_{\text{diff}} = 0.1$) & {-0.078} & \textbf{0.237} & \textbf{0.938} & \textbf{0.141} & {-0.013} & \textbf{0.266} & \textbf{0.927} & \textbf{0.168}\\
         Diffused Censored GP (fixed $\pi_{\text{diff}} = 0.2$ $\leftarrow$ true) & \textbf{-0.084} & {0.238} & \textbf{0.938} & {0.143} & \textbf{-0.019} & {0.268} & {0.926} & {0.172}\\
         Diffused Censored GP (fixed $\pi_{\text{diff}} = 0.3$) & {-0.081} & {0.242} & {0.936} & {0.150} & {-0.016} & {0.276} & {0.923} & {0.183}\\
         Diffused Censored GP (fixed $\pi_{\text{diff}} = 0.4$) & {-0.076} & {0.243} & {0.936} & {0.153} & {-0.011} & {0.277} & {0.923} & {0.186}\\
         Diffused Censored GP (fixed $\pi_{\text{diff}} = 0.5$) & {-0.066} & {0.246} & {0.935} & {0.157} & {-0.001} & {0.279} & {0.921} & {0.190}\\
         Diffused Censored GP (fixed $\pi_{\text{diff}} = 0.6$) & {-0.052} & {0.249} & {0.932} & {0.164} & {0.013} & {0.283} & {0.919} & {0.196}\\
         Diffused Censored GP (fixed $\pi_{\text{diff}} = 0.7$) & {-0.034} & {0.254} & {0.930} & {0.171} & {0.031} & {0.288} & {0.916} & {0.203}\\
         \midrule
         Diffused Censored GP (learned $\pi_{\text{diff}}$) & {-0.084} & {0.240} & {0.937} & {0.146} & {-0.018} & {0.272} & {0.925} & {0.177}\\
        \bottomrule
    \end{tabular}
    \caption{Sensitivity analysis results for model misspecification with respect to the diffusion hyper-parameter $\pi_{\text{diff}}$ (sink node probability).}
    %Includes: paper - demo\_fakeSTdata9\_STkernel.ipynb}
    \label{tab:fakeSTdata_wrong_pi}
\end{table*}

Based on the Dataset-ST1 (for which the true demand is well specified) we performed a sensitivity analysis with respect to model misspecification regarding the diffusion lengthscale $\ell_{\text{diff}}$ (Table~\ref{tab:fakeSTdata_wrong_ell}), as well as for the sink node probability $\pi_{\text{diff}}$ (Table~\ref{tab:fakeSTdata_wrong_pi}). The goal of this analysis is to understand the impact of using an incorrect value of these diffusion hyper-parameters, as it may happen when relying on domain knowledge to set these values, on the ability to model the true demand of the proposed approach. As expected, the results in Table~\ref{tab:fakeSTdata_wrong_ell} show that, as the value of $\ell_{\text{diff}}$ deviates from the true value of 0.1 (that was used to generate the data), the performance deteriorates. However, the performance decay is not very abrupt, and even if this parameter is wrongly set by a factor of 2, the results remain largely unchanged. If we consider, for example, the problem of inferring the true demand for bike-sharing systems, the value of $\ell_{\text{diff}}$ can be interpreted as the customers' willingness to walk to the next hub to pick up a bicycle in case the nearest hub is out of supply. Domain knowledge (and common sense) allows the modeler to set this value fairly confidently, and knowing that misspecification up to a factor of 2 does not seem to have a noticeable impact on performance is a strong indicator of the robustness of the proposed approach. Furthermore, note that learning $\ell_{\text{diff}}$ is always an option, although this corresponds to a complex optimization problem. Lastly, it should be noted that for all the values of $\ell_{\text{diff}}$ tested in Table~\ref{tab:fakeSTdata_wrong_ell}, the results of the proposed approach are still better than those obtained for the Censored GP, even when the value of $\ell_{\text{diff}}$ is misspecified by one order of magnitude. 

Similarly, Table~\ref{tab:fakeSTdata_wrong_pi} shows the impact of misspecification of the sink node probability $\pi_{\text{diff}}$. Interestingly, if the value of $\pi_{\text{diff}}$ is set higher than the true value used to generate the data ($0.2$), the performance deteriorates (as one would expect), however, if the value of $\pi_{\text{diff}}$ is underestimated, then the performance of the proposed approach obtained was even (marginally) better. We hypothesize that this is due to the model more aggressively trying to use diffusion to explain the observed demand data, which in this case actually helps it to be less susceptible to observation noise ($\sigma^2$). Importantly, as with the sensitivity analysis with respect $\ell_{\text{diff}}$, the results in Table~\ref{tab:fakeSTdata_wrong_pi} show that for all the values of $\pi_{\text{diff}}$ tested, the results of the proposed approach are still significantly better than those obtained for the Censored GP.

\subsection{Learning $\pi_{\textnormal{diff}}$}
\label{appendix:learning_pi}

In this subsection, we study the possibility of the proposed approach being able to learn the hyper-parameter $\pi_\text{sink} \in \mathbb{R}$, which controls the likelihood of customers simply giving up and deciding not to consume any alternative product when faced with the lack of supply for their intended product. As explained in Section~\ref{subsec:diff_param_learning}, we propose to achieve this via type-II maximum likelihood by maximizing the ELBO in Eq.~\ref{eq:elbo} using gradient descent. Table~\ref{tab:fakeSTdata_learning_pi} shows the obtained results for different initialization of $\pi_{\text{diff}}$. As the results show, initialization plays a role. This is unsurprising since hyper-parameter learning in GPs is well known to be a challenging non-convex optimization problem. Nevertheless, the proposed approach was able to achieve results close to the results of the proposed approach with $\pi_{\text{diff}}$ fixed to the true value used to generate the data ($\pi_{\text{diff}}=0.2$) when initialized with 0.2 and 0.5, with the learned value converging in both cases to $\pi_{\text{diff}} \approx 0.236$. When initialized with $\pi_{\text{diff}}=0$, the learned value remained 0, which, in this case, actually leads to even better results. This is also consistent with the results in Table~\ref{tab:fakeSTdata_wrong_pi}. 

\begin{table*}[!th]
    \centering
    \small
    \begin{tabular}{l | rrrr | rrrr}
        \toprule
	& \multicolumn{4}{c}{\textbf{Trainset}} & \multicolumn{4}{|c}{\textbf{Testset}} \\
	\textbf{Model} & \textbf{NLPD} & \textbf{RMSE}  & $\textbf{R}^2$ & $\textbf{RMSE funct.}$ & \textbf{NLPD} & \textbf{RMSE} & $\textbf{R}^2$ & $\textbf{RMSE funct.}$ \\ 
        \midrule
         %True GP (Oracle) & -0.094 & 0.239 & 0.937 & 0.144 & -0.031 & 0.273 & 0.924 & 0.178\\
         %(Non-Censored) GP & 0.220 & 0.325 & 0.893 & 0.286 & 0.327 & 0.368 & 0.869 & 0.313\\
         Censored GP & 0.043 & 0.276 & 0.915 & 0.207 & 0.111 & 0.313 & 0.900 & 0.239\\
         DCGP ($\ell_{\text{diff}} = 0.1$, $\pi_{\text{diff}} = 0.2$) & \textbf{-0.084} & \textbf{0.238} & \textbf{0.938} & \textbf{0.143} & \textbf{-0.019} & \textbf{0.268} & \textbf{0.926} & \textbf{0.172}\\
         %DCGP ($\ell_{\text{diff}} = $learned, $\pi_{\text{diff}} = 0.2$) & {-0.015} & {0.251} & {0.931} & {0.167} & {0.047} & {0.281} & {0.920} & {0.191}\\
        \midrule
         DCGP ($\ell_{\text{diff}} = 0.1$, $\pi_{\text{diff}} = $learned; init=0.0) & {-0.066} & {0.239} & {0.936} & {0.142} & {-0.002} & {0.267} & {0.926} & {0.169}\\
         DCGP ($\ell_{\text{diff}} = 0.1$, $\pi_{\text{diff}} = $learned; init=0.2) & {-0.084} & {0.240} & {0.937} & {0.146} & {-0.018} & {0.272} & {0.925} & {0.177}\\
         DCGP ($\ell_{\text{diff}} = 0.1$, $\pi_{\text{diff}} = $learned; init=0.5) & {-0.084} & {0.240} & {0.937} & {0.146} & {-0.018} & {0.272} & {0.925} & {0.178}\\
        \bottomrule
    \end{tabular}
    \caption{Experiments results for learning the hyper-parameter $\pi_{\text{diff}}$ for the Dataset-ST1. ``RMSE funct.'' further measures the RMSE to the true underlying (noiseless) function.}
    %Includes: paper - demo\_fakeSTdata9\_STkernel.ipynb}
    \label{tab:fakeSTdata_learning_pi}
\end{table*}

\subsection{Impact of trainset size for sales data}
\label{appendix:sales_splits}

\begin{table*}[!th]
    \centering
    \small
    \begin{tabular}{l | rrr | rrr}
        \toprule
	& \multicolumn{3}{c}{\textbf{Trainset}} & \multicolumn{3}{|c}{\textbf{Testset}} \\
         \textbf{Model} & \textbf{NLPD} & \textbf{RMSE}  & $\textbf{R}^2$ & \textbf{NLPD} & \textbf{RMSE} & $\textbf{R}^2$ \\ 
%        \midrule
%         True GP (Oracle) & 1.890 & 0.613 & 0.622 & 2.156 & 0.709 & 0.506\\
%         GP & 2.164 & 0.696 & 0.511 & 2.334 & 0.774 & 0.412\\
%         Censored GP & 2.182 & 0.685 & 0.527 & 2.344 & 0.770 & 0.418\\
%         Diffused Censored GP (fixed) & \textbf{1.975} & \textbf{0.615} & \textbf{0.622} & \textbf{2.198} & \textbf{0.697} & \textbf{0.525}\\
%         Diffused Censored GP (fixed) & \textbf{1.972} & \textbf{0.611} & \textbf{0.627} & \textbf{2.194} & \textbf{0.696} & \textbf{0.525}\\
%         Diffused Censored GP (learned) & {2.058} & {0.639} & {0.592} & {2.261} & {0.727} & {0.483}\\
%         Diffused Censored GP (learned) & {2.011} & {0.623} & {0.612} & {2.276} & {0.713} & {0.502}\\
        \midrule
         True GP (Oracle) & 1.890 & 0.613 & 0.622 & 2.156 & 0.709 & 0.506\\
         (Non-Censored) GP & 2.117 & 0.695 & 0.512 & 2.229 & 0.737 & 0.467\\
         Censored GP & 2.176 & 0.718 & 0.477 & 2.189 & 0.720 & 0.491\\
         %Diffused Censored GP (fixed) & \textbf{1.664} & \textbf{0.606} & \textbf{0.634} & \textbf{1.996} & \textbf{0.664} & \textbf{0.566}\\
         Diffused Censored GP (fixed) & \textbf{1.674} & \textbf{0.606} & \textbf{0.633} & \textbf{1.988} & \textbf{0.664} & \textbf{0.566}\\
         %Diffused Censored GP (learned) & {1.819} & {0.626} & {0.609} & {2.124} & {0.689} & {0.534}\\
         Diffused Censored GP (learned) & {1.761} & {0.625} & {0.611} & {2.025} & {0.675} & {0.551}\\
        \bottomrule
    \end{tabular}
    \caption{Experimental results obtained for supermarket sales dataset using a 70/30\% train/test data split.}
    %paper - demo\_salesData7030\_indepKernel\_throwExcess.ipynb}
    \label{tab:sales7030}
\end{table*}

\begin{table*}[!th]
    \centering
    \small
    \begin{tabular}{l | rrr | rrr}
        \toprule
	& \multicolumn{3}{c}{\textbf{Trainset}} & \multicolumn{3}{|c}{\textbf{Testset}} \\
         \textbf{Model} & \textbf{NLPD} & \textbf{RMSE}  & $\textbf{R}^2$ & \textbf{NLPD} & \textbf{RMSE} & $\textbf{R}^2$ \\ 
        \midrule
%         True GP (Oracle) & 1.770 & 0.565 & 0.700 & 2.284 & 0.762 & 0.381\\
%         GP & 2.321 & 0.697 & 0.539 & 2.678 & 0.844 & 0.239\\
%         Censored GP & 2.256 & 0.733 & 0.496 & 2.537 & 0.811 & 0.297\\
%         Diffused Censored GP (fixed) & \textbf{1.777} & \textbf{0.632} & \textbf{0.631} & \textbf{2.430} & 0.778 & 0.358\\
%         Diffused Censored GP (learned) & {2.240} & {0.642} & {0.619} & {2.655} & \textbf{0.770} & \textbf{0.372}\\
%        \midrule
         True GP (Oracle) & 1.770 & 0.565 & 0.700 & 2.284 & 0.762 & 0.381\\
         (Non-Censored) GP & 2.309 & 0.694 & 0.542 & 2.605 & 0.815 & 0.291\\
         Censored GP & 2.149 & 0.717 & 0.517 & 2.359 & 0.764 & 0.379\\
         %Diffused Censored GP (fixed) & \textbf{1.718} & \textbf{0.627} & \textbf{0.636} & \textbf{2.275} & 0.729 & 0.435\\
         Diffused Censored GP (fixed) & \textbf{1.726} & \textbf{0.627} & \textbf{0.636} & \textbf{2.291} & 0.730 & 0.434\\
         %Diffused Censored GP (learned) & {2.362} & {0.651} & {0.607} & {2.643} & \textbf{0.737} & \textbf{0.423}\\
         Diffused Censored GP (learned) & {2.008} & {0.667} & {0.586} & {2.513} & \textbf{0.738} & \textbf{0.421}\\
        \bottomrule
    \end{tabular}
    \caption{Experimental results obtained for supermarket sales dataset using a 50/50\% train/test data split.}
    %paper - demo\_salesData5050\_indepKernel\_throwExcess.ipynb}
    \label{tab:sales5050}
\end{table*}

\begin{table*}[!th]
    \centering
    \small
    \begin{tabular}{l | rrr | rrr}
        \toprule
	& \multicolumn{3}{c}{\textbf{Trainset}} & \multicolumn{3}{|c}{\textbf{Testset}} \\
         \textbf{Model} & \textbf{NLPD} & \textbf{RMSE}  & $\textbf{R}^2$ & \textbf{NLPD} & \textbf{RMSE} & $\textbf{R}^2$ \\ 
%        \midrule
%         True GP (Oracle) (Gaussian) & 2.250 & 1.676 & 0.947 & 3.296 & 6.751 & 0.341\\
%         True GP (Oracle) (Poisson) & 1.988 & 1.653 & 0.948 & 2.658 & 6.574 & 0.375\\
%         GP (Gaussian) & 2.701 & 3.686 & 0.745 & 3.357 & 7.099 & 0.273\\
%         GP (Poisson) & 2.275 & 3.515 & 0.768 & 2.711 & 6.984 & 0.294\\
%         Censored GP (Gaussian) & 2.572 & 3.550 & 0.760 & 3.317 & 6.813 & 0.328\\
%         Censored GP (Poisson) & 2.191 & 4.058 & 0.691 & 2.665 & 6.889 & 0.316\\
%         Diffused Censored GP (fixed) (Gaussian) & 2.224 & {2.747} & {0.859} & 3.316 & 6.878 & 0.314\\
%         Diffused Censored GP (fixed) (Poisson) & {2.099} & 3.064 & 0.826 & {2.656} & {6.698} & {0.353}\\
%         Diffused Censored GP (learned) (Gaussian) & 2.201 & \textbf{2.590} & \textbf{0.873} & 3.304 & 6.800 & 0.330\\
%         Diffused Censored GP (learned) (Poisson) & \textbf{2.095} & 3.020 & 0.831 & \textbf{2.652} & \textbf{6.659} & \textbf{0.360}\\
        \midrule
         True GP (Oracle) & 2.250 & 1.676 & 0.947 & 3.296 & 6.751 & 0.341\\
         %True GP (Oracle) (Poisson) & 1.988 & 1.653 & 0.948 & 2.658 & 6.574 & 0.375\\
         (Non-Censored) GP & 2.668 & 3.535 & 0.768 & 3.346 & 7.049 & 0.284\\
         %GP (Poisson) & 2.275 & 3.515 & 0.768 & 2.711 & 6.984 & 0.294\\
         Censored GP & 2.615 & 3.954 & 0.704 & 3.309 & 6.804 & 0.331\\
         %Censored GP (Poisson) & 2.191 & 4.058 & 0.691 & 2.665 & 6.889 & 0.316\\
         %Diffused Censored GP (fixed) (Gaussian) & 2.136 & {2.656} & {0.867} & 3.305 & 6.824 & 0.326\\
         Diffused Censored GP (fixed) & 2.141 & {2.776} & {0.855} & 3.306 & 6.839 & 0.322\\
         %Diffused Censored GP (fixed) (Poisson) & {2.099} & 3.064 & 0.826 & {2.656} & {6.698} & {0.353}\\
         %Diffused Censored GP (learned) (Gaussian) & 2.149 & \textbf{2.778} & \textbf{0.854} & 3.296 & 6.777 & 0.335\\
         Diffused Censored GP (learned) & 2.165 & \textbf{3.018} & \textbf{0.829} & 3.304 & 6.800 & 0.331\\
         %Diffused Censored GP (learned) (Poisson) & \textbf{2.095} & 3.020 & 0.831 & \textbf{2.652} & \textbf{6.659} & \textbf{0.360}\\
        \bottomrule
    \end{tabular}
    \caption{Experimental results obtained for a version of the bike-sharing dataset considering only two hubs.}
    %paper - demo\_bikeData7030\_STkernel\_bugfix\_v2\_throwExcess.ipynb}
    \label{tab:bike_2stations}
\end{table*}

The supermarket sales dataset used in Section~\ref{subsec:sales_main} is rather small and, for smaller train set sizes, signs of overfitting quickly start to emerge. Therefore, this provides a good opportunity to analyze the impact of train set size on the performance of the different methods. Besides the 90/10\% train/test split considered in the main body of the paper (Table~\ref{tab:sales9010}), we also considered 70/30\% and 50/50\% train/test splits - shown in Tables~\ref{tab:sales7030} and \ref{tab:sales5050}, respectively. Despite the clear sign of overfitting, whereby the train set performance improved as the train set size decreased while the test set performance became worse, the results in Tables~\ref{tab:sales7030} and \ref{tab:sales5050} show that the proposed approach still outperforms all the other baseline approaches.

\subsection{Bike sharing with only two hubs}
\label{appendix:bike_2stations}

Besides the experiments with 4 hubs, we also performed experiments using a simplified version using only 2 hubs of the bike-sharing dataset. Table~\ref{tab:bike_2stations} shows the obtained results. The latter remain consistent with the findings from the 4-hub experiment, thus further providing further evidence of the advantages of the proposed approach across a wide broad of scenarios and application domains.  

\vfill

\end{document}